\documentclass{article}

     \PassOptionsToPackage{numbers, compress}{natbib}

\usepackage[preprint]{neurips_data_2023}

\usepackage[utf8]{inputenc} 
\usepackage[T1]{fontenc}    
\usepackage{hyperref}       
\usepackage{url}            
\usepackage{booktabs}       
\usepackage{amsfonts}       
\usepackage{nicefrac}       
\usepackage{microtype}      
\usepackage{xcolor}         
\usepackage{graphicx}
\usepackage{comment}
\usepackage{enumitem}
\usepackage{float}
\usepackage{algpseudocode,algorithm}
\usepackage{minitoc}

\title{Comparing Reinforcement Learning and Human Learning using the Game of Hidden Rules}

\author{Eric Pulick$^{1*}$, Vladimir Menkov$^2$\\ \textbf{Yonatan Mintz$^1$, Paul Kantor$^1$, Vicki M.~Bier$^1$}\\
 $^1$University of Wisconsin - Madison, $^2$Rutgers University\\
\{pulick, ymintz, pkantor, vmbier\}@wisc.edu, vmenkov@gmail.com
}

\begin{document}
\doparttoc
\faketableofcontents
\maketitle

\begin{abstract}
Reliable real-world deployment of reinforcement learning (RL) methods requires a nuanced understanding of their strengths and weaknesses and how they compare to those of humans. Human-machine systems are becoming more prevalent and the design of these systems relies on a task-oriented understanding of both human learning (HL) and RL. Thus, an important line of research is characterizing how the structure of a learning task affects learning performance. While increasingly complex benchmark environments have led to improved RL capabilities, such environments are difficult to use for the dedicated study of task structure. To address this challenge we present a learning environment built to support rigorous study of the impact of task structure on HL and RL. We demonstrate the environment's utility for such study through example experiments in task structure that show performance differences between humans and RL algorithms.
\end{abstract}

\section{Introduction}
Reinforcement learning (RL) \citep{sutton_reinforcement_2018} benchmarks often come directly from human benchmarks (e.g., games) or are designed to mimic complex reasoning tasks a human might encounter. These increasingly complex benchmark environments have been used to improve the capability of RL algorithms, leading to impressive achievements in domains such as board games \citep{samuel_studies_1959,tesauro_td-gammon_1994,silver_mastering_2016,silver_mastering_2017,silver_general_2018,schrittwieser_mastering_2020} and video games \citep{mnih_human-level_2015,jaderberg_human-level_2019,vinyals_grandmaster_2019,openai_dota_2019}. Unfortunately, while the difficulty of such environments motivates greater RL capabilities, their complexity often makes them ill-suited for the rigorous study of task structure; that is, how the logical structure of a learning task affects learning performance. A task-oriented understanding of RL methods' strengths and weaknesses remains an important gap in the informed deployment of RL tools. In particular, such an understanding would allow decision-makers to better relate findings from benchmark environments to the specific attributes of their own problem settings. 

Increasingly, practitioners must also decide how to integrate RL tools alongside humans. A granular, task-oriented understanding of \emph{both} human learning (HL) and RL is necessary to design systems where humans and algorithms best complement one another. Changes in task attributes that make learning easier for either HL or RL but harder for the other may suggest cases where human-machine learning pairs can be effective \citep{hadfield-menell_cooperative_2016,bansal_beyond_2019,bansal_is_2021}. Direct HL-RL comparisons can also identify helpful human priors or heuristics for future algorithm development. An important step in this line of research is the development of learning environments that support the study of task structure for both HL and RL.

In this paper we study learning tasks within the classical RL setting, where agents learn through sequential interaction with an uncertain environment. Specifically, we consider a new benchmark environment that can be used to systematically assess how the logical structure of learning tasks affects the performance of HL and RL (accounting for hyperparameter selections, feature representations, etc.). Our environment has several advantages over existing environments for the dedicated study of task structure. Existing environments vary in multiple ways, making direct performance comparisons challenging to interpret. Likewise, due to their complexity, it is difficult to use variations of existing environments to create generalizable findings about the impact of task structure. For example, while the objective of chess is clear, the underlying learning task is a complex composition of the board structure, game mechanics, and adversarial dynamics. An experiment might explore variations to this learning task by modifying the movement patterns of different pieces \citep{tomasev_reimagining_2022}. However, without a clear understanding of how the existing elements of chess contribute to learning performance, it is unclear how to incorporate these results into a generalized understanding of the impact of task structure.

\textbf{Contributions } We present a novel RL environment, called the Game of Hidden Rules (GOHR), that allows researchers to rigorously investigate the impact of task structure on learning performance. This extends preliminary work on this environment \citep{pulick_game_2022}, as discussed in Appendix~\ref{app:prev_work}. The GOHR complements existing learning environments and distinguishes itself as a useful tool for the study of task structure in three substantive ways. First, each hidden rule encodes a clearly defined logical pattern as the learning objective, allowing researchers to draw systematic distinctions between learning tasks. Second, GOHR's rule syntax allows for fine variations in task definition, enabling experiments that study controlled differences in learning tasks. Third, GOHR's rule syntax introduces a vast space of hidden rules for study, ranging from trivial to complex, providing an appropriate starting point for the study of task structure. We demonstrate the use of the GOHR through two example experiments in task structure that compare human learners to sample RL algorithms.

\section{Related work}\label{sec:related_work}

\textbf{RL environments }
As noted, recent improvements in RL algorithms can be credited to an explosion in simulation-based benchmarking environments. Tools such as the Arcade Learning Environment \citep{bellemare_arcade_2013}, openAI gym \citep{brockman_openai_2016}, modern video games \citep{kempka_vizdoom_2016,vinyals_starcraft_2017,guss_minerl_2019,wei_honor_2022}, and procedurally generated environments \citep{cobbe_leveraging_2020,juliani_obstacle_2019,kuttler_nethack_2020,samvelyan_minihack_2021} have spurred RL development through increasingly complex and realistic problem settings. For instance, procedurally generated environments motivate more robust learning algorithms that can better handle variations to the environment. Other environments highlight cooperative or adversarial challenges unique to multi-agent settings \citep{suarez_neural_2021}, further expanding the breadth of tasks RL algorithms face. Collectively, these environments raise expectations for state-of-the-art RL algorithm performance. However, their emphasis on challenging high-end capabilities of algorithms often makes them difficult starting points for fundamental studies into the impact of task structure. The GOHR is intended to address this unmet need in the space of RL environments by allowing researchers to design precise experiments investigating the impact of task structure on learning.

\textbf{Analysis of RL performance }
\citet{islam_reproducibility_2017} and \citet{henderson_deep_2018} initiated important efforts to assess the reproducibility of RL performance and look more deeply at the effects of different internal design choices (e.g., network architectures) on performance \citep{wang_benchmarking_2019,andrychowicz_what_2020,andrychowicz_what_2021}. Such studies offer a great deal of practical insight related to algorithm design choices, but do not generally clarify task-oriented differences in tested benchmark environments. Among broad efforts to compare algorithm performance across benchmarks, the \verb|bsuite|, introduced by \citet{osband_behaviour_2020}, is most closely aligned with our line of research. The \verb|bsuite| identifies high-level desired characteristics of effective learning agents (generalization, exploration, and handling of long-term consequences) and gathers a set of benchmark environments to assess the performance of different algorithms against these characteristics. While the \verb|bsuite| concentrates on higher-level performance characteristics, the GOHR provides a complementary testbed focused specifically on the logical structure of the task to be learned and its impact on learning performance. Both approaches mark important steps toward a more nuanced understanding of RL algorithms.

\textbf{Comparisons of humans to machine learning }
There is a growing number of studies comparing the performance of machine learning (ML) to humans on particular real-world tasks, such as medical imaging \citep{liu_comparison_2019}. Similarly, the RL benchmarking literature often measures algorithm performance against human-level performance. These analyses provide valuable reference points for the performance of humans and state-of-the-art ML algorithms on particular tasks, but they do not clarify fundamental questions regarding the impact of task structure on learning performance. To our knowledge, there is little research that addresses rigorous ML/HL comparisons \textit{with respect to task structure}. We believe a primary reason for the gap in literature investigating task structure, particularly within RL, is the present lack of environments capable of supporting small, precise, and interpretable changes to tasks. As noted by \citet{hernandez-orallo_evaluation_2017} and \citet{burnell_rethink_2023}, more granular evaluation metrics are needed to properly interpret ML capabilities; this need for granularity holds when ML capabilities are compared rigorously to human performance \citep{cowley_framework_2022}. Deeper investigations into HL/RL responses to task structure may give important insight into algorithm design for more ambitious benchmarks like the Abstract Reasoning Corpus (ARC) \citep{chollet_measure_2019}. With respect to human-ML comparisons, most similar to our work is that of \citet{kuhl_human_2022}, which examines a range of pattern recognition tasks in a supervised learning setting. As in our work, they present a curated set of tasks and demonstrate differences between the performance of human players and various algorithms.

\section{Game of Hidden Rules}\label{sec:gohr}
This overview of the GOHR closely follows \cite{pulick_game_2022}. Additional details can be found in Appendix~\ref{app:gohr}. Comprehensive documentation and the complete toolset are available at our \href{http://sapir.psych.wisc.edu:7150/w2020/captive.html}{public site}.

\begin{figure}[h]
    \centering
    \begin{minipage}{.43\textwidth}
        \centering
        \includegraphics[width=.7\textwidth]{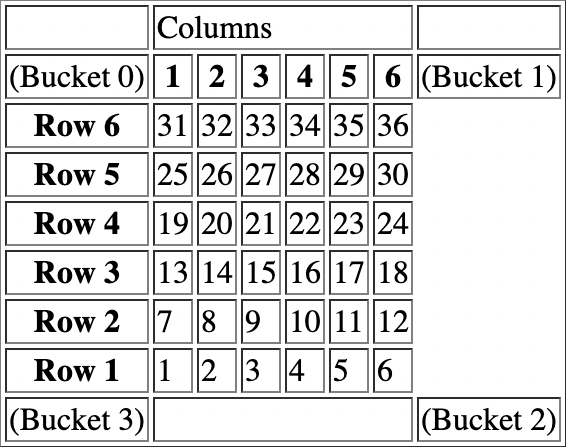}
    \end{minipage}
    \hspace{.5cm}
    \begin{minipage}{0.43\textwidth}
        \centering
        \includegraphics[width=.6\textwidth]{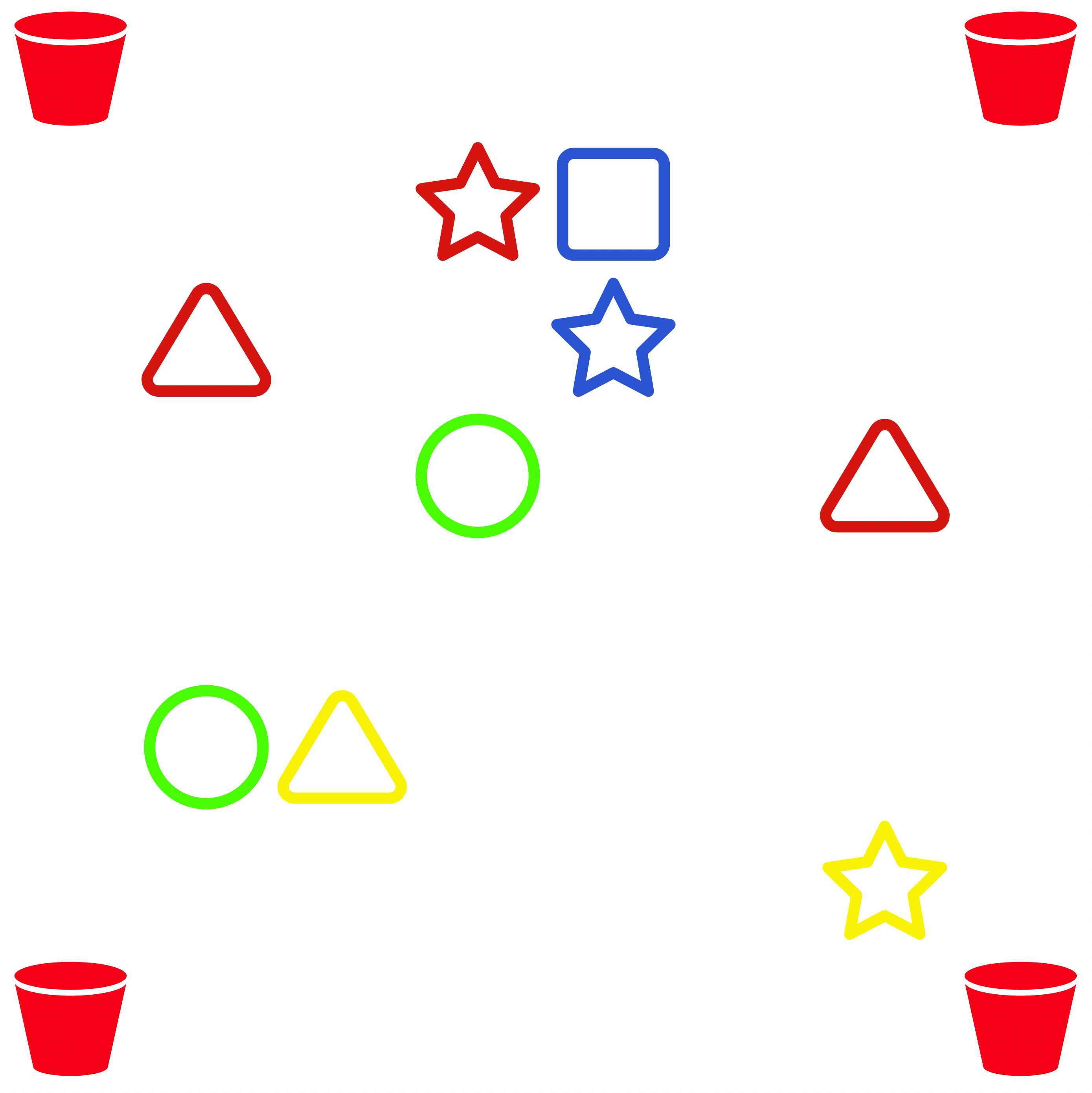}
    \end{minipage}
    \caption{Game board diagram (left) and a sample board with four shapes and colors (right).}
    \label{fig:boards}
\end{figure}

\textbf{Game board }
The GOHR is played on a $6\times 6$ grid-style board using game pieces of varying shapes and colors. At the beginning of an episode, the game engine populates the board with game pieces; the player's goal is to clear the board of game pieces by placing them, one at a time, into any of the four buckets located at the corners of the board. Figure~\ref{fig:boards} provides a diagram of the board where all individual board cells, rows, columns, and buckets are given numeric labels along with a sample board as a human player would see it. Note that the collection of shapes and colors used for a given experiment is entirely configurable by the researcher, with a default set of four shapes and four colors. If desired, researchers can construct exact board layouts in advance of play (to be seen randomly or in a specified order), otherwise pieces are generated randomly per a set of input parameters. This flexibility allows the experimenter to design experiments addressing the learning curricula itself (e.g., to determine if seeing particular game pieces affects the performance of the learner for a given rule). Additional details are provided in Appendix~\ref{app:gohr_boards}. 

\textbf{Hidden rules }
A hidden rule, known to the researcher but not to the player, determines which pieces may be placed into which buckets. For example, a rule might allow pieces into certain buckets based on their shape, color, or position on the board. When the player makes a move (i.e., tries placing a particular game piece into a bucket), they receive immediate feedback on whether the move is allowed; if the move is allowed, the piece is removed from play, otherwise the piece returns to its original place on the board. Hidden rules are constructed from one or more \textbf{rule lines}, each of which is built from one or more \textbf{atoms}. For instance, a two-line rule with five atoms might look like:

{\centering
    (atom 1) (atom 2) (atom 3)\\
    (atom 4) (atom 5)
\par}

Only one rule line is \textbf{active} at a time;  this active line determines the current \textbf{rule state} (how game pieces may be placed into buckets for the player's current move). In the example above, the rule state is formed by the contents of either atoms 1, 2, and 3 or atoms 4 and 5, depending on which line is active. Each atom maps a set of game pieces to a set of permitted buckets and is defined as follows:

{\centering
    (count, shapes, colors, positions, buckets)
\par}

Any game pieces matching the \textit{shapes}, \textit{colors}, and \textit{positions} specified in the atom are accepted in any element of its set of \textit{buckets}. The \textit{count} parameter defines the number of successful moves for which the atom remains valid and is used in rules where the rule state changes during play. Multiple values can be listed for each non-count field and are grouped in brackets. A simple example where stars and triangles always go in the top-left bucket (0) while circles and squares always go in the bottom-right bucket (2), regardless of their color or position, can be expressed with two atoms on one rule line:

{\centering (*, [star,triangle], *, *, 0) (*, [circle,square], *, *, 2)\par}

The $*$ character is a wildcard. For the count field, $*$ means the atom is always valid; for shape, color, or position, it means any value for that feature is permissible. A set of helpful rule examples, illustrating the expressiveness of the rule syntax, can be found in Appendix~\ref{app:gohr_rules}. Broadly, rules within the GOHR can be divided into two categories: \textbf{stationary} and \textbf{non-stationary}. Stationary rules are those in which the rule state does not change during game play. In such rules, whether a move is permitted does not depend on the state of the board or past actions made by the player. These rules can be used to evaluate a player's ability to learn strictly feature-based patterns. The example noted above is stationary; the board state and move history do not impact where game pieces are permitted. In contrast, non-stationary rules are those in which the rule state changes during play, meaning that permitted moves will depend on the state of the board or past successful moves the player has made. Non-stationary rules embed temporal components into the logical pattern the player must learn. The rules presented in Appendix~\ref{app:gohr_rules} further describe the mechanics available to researchers in creating non-stationary rules. Importantly, we note that non-stationary rules update the rule state only after successful moves. The rule state is unchanged after an incorrect move.

\section{Example rules}\label{sec:conceptual_framework}
In this section, we introduce the rules used in our experiments. Our first experiment explores a few \textbf{stationary} and \textbf{non-stationary} learning task structures expressible within the GOHR. We consider the following rules (see Appendix~\ref{app:rule_syntax} for related rule syntax and our \href{http://sapir.psych.wisc.edu:7150/w2020/launch/launch-rules-cgs.jsp}{public site} to play these rules):
\begin{enumerate}[topsep=0pt,itemsep=-0.5ex,leftmargin=.75cm]
    \item[--]Shape Match (\textbf{SM}) - Each of the four shapes is mapped uniquely to a bucket (i.e., stars go in bucket 0, triangles in bucket 1, squares in bucket 2, circles in bucket 3).
    \item[--]Quadrant Nearby (\textbf{QN}) - Pieces in each board quadrant are mapped uniquely to the bucket nearest to that quadrant.
    \item[--]Bottom-left then Top-right (\textbf{BLTR}) - The rule state alternates between allowing a piece in the bottom-left bucket (3) and allowing a piece in the top-right bucket (1).
    \item[--]Clockwise (\textbf{CW}) - The first piece must be placed in the top-left bucket (0) and each subsequent piece must be placed in the next clockwise bucket (i.e., the pattern follows buckets 0-1-2-3).
\end{enumerate}

Note that these rules are equivalently difficult at random; in each, a random policy will always have a $\nicefrac{1}{4}$ chance of making a correct move, regardless of the board state or past moves. The rules, however, task the player with learning patterns with different logical structures. Shape Match and Quadrant Nearby are stationary and use a single game piece feature (shape or position). Bottom-left then Top-right and Clockwise are non-stationary, encoding sequences of length 2 and 4, respectively. Our aim with this experiment is to demonstrate how HL and RL may respond differently to specific variations in task structure. Identifying how specific logical structures within learning tasks affect performance for RL or HL players will better inform the analysis of more complex environments, where learning tasks may be compositions of fundamental logical structures expressible in the GOHR.

Our second experiment addresses a characteristic of learning tasks that we call \textbf{rule generality}. Broadly, rule generality reflects that multiple policies may be effective for a particular rule. More rigorously, let a player's \textbf{move policy} be the policy by which they generate their moves. Such a policy is \textbf{sufficient} if applying this policy to any possible board state and sequence of past actions yields error-free play. A given move policy may be sufficient to many rules and rules may permit many sufficient policies. For example, consider the stationary rule where red and blue game pieces are permitted in buckets 0 and 1 while green and yellow game pieces are permitted in buckets 2 and 3. A sufficient policy could be to select game pieces and associated actions in the following order:

{\centering
    red $\rightarrow$ bucket 0,$\quad$ blue $\rightarrow$ bucket 1,$\quad$ green $\rightarrow$ bucket 2,$\quad$ yellow $\rightarrow$ bucket 3
\par}

Any policy relying on these same color-to-bucket mappings would also be sufficient, regardless of the order in which it selects pieces. Further consider arbitrary rules $\mathcal{A}$ and $\mathcal{B}$. With respect to generality, $\mathcal{A}$ \textbf{properly dominates} $\mathcal{B}$ if any sufficient policy for $\mathcal{B}$ is sufficient for $\mathcal{A}$ and there exists a sufficient policy for $\mathcal{A}$ that is not sufficient for $\mathcal{B}$. We refer to $\mathcal{A}$ as more general than $\mathcal{B}$ (denoted $\mathcal{A} \succ \mathcal{B}$). To study the response of players to increasing rule generality, we consider the following variations of the rules given in our first experiment (see Appendix~\ref{app:rule_syntax} for expression in GOHR rule syntax):\footnote{We also studied a set of color-based rules (CM, CM1F, CM2O) mirroring the structure of shape rules SM, SM1F, and SM2O. Results were identical to shape rules for RL players and very similar for humans, see Appendix~\ref{app:color_rules}.}
\begin{enumerate}[topsep=0pt,itemsep=-0.5ex,leftmargin=.75cm]
    \item[--]Shape Match 1 Free (\textbf{SM1F}) - Shape Match modified so one shape can be placed in any bucket.
    \item[--]Shape Match 2 Options (\textbf{SM2O}) - Similar to Shape Match, except each of the four shapes is mapped to two buckets rather than one.
    \item[--]Quadrant Nearby 2 Free (\textbf{QN2F}) - Identical to Quadrant Nearby, except that pieces in two of the quadrants can be placed in any bucket.
    \item[--]Bottom then Top (\textbf{BT}) - The rule state alternates between allowing a piece in either of the bottom two buckets (2,3) and allowing a piece in either of the top two buckets (0,1).
    \item[--]Clockwise Alternating Free (\textbf{CWAF}) - Same as Clockwise, but every other move is a free move (i.e., any piece will be accepted in any bucket), per the repeating pattern 0-*-2-*.
    \item[--]Clockwise 2 Free (\textbf{CW2F}) - Same as Clockwise, except the last two moves in the 0-1-2-3 bucket pattern are free moves, i.e., following the repeating pattern 0-1-*-*.
\end{enumerate}

Each rule is constructed to properly dominate a corresponding `base rule' from the first experiment (e.g., CWAF $\succ$ CW, CW2F $\succ$ CW). When we provide two rule variations of a base rule (e.g., CWAF, CW2F), neither is more general than the other. Our aim with this experiment is to study the responses of human players and RL algorithms to increasing generality by comparing performance on more general rule variations to their respective base rules.  

\section{Experimental setup}\label{sec:experiments}
We describe the human and RL participants in our experiments, experimental procedures, performance metrics, and statistical comparison of our results. Portions of this section closely follow \citep{pulick_game_2022}.

\textbf{Human participants }
Human participants in our GOHR experiments came from the Amazon Mechanical Turk platform \citep{buhrmester_amazons_2016}, a popular tool for crowdsourcing tasks and research. Each player received a brief set of instructions about the mechanics of the GOHR and subsequently played 3-7 episodes of the same rule as part of their participation in the experiment. Players were selected such that they had no prior exposure to the GOHR. Approximately 25 participants were assigned to each rule listed in Section~\ref{sec:conceptual_framework}. In each episode, players received boards randomly populated with 8 or 9 pieces, depending on the rule. Additional information regarding experimental flow, subject counts, payments, and board generation parameters can be found in Appendix~\ref{app:human_exp}.

\textbf{RL algorithms }
To describe our two sample algorithms, we model the GOHR as a Markov Decision Process (MDP). The state observed by the player at time $t$, $S_t \in \mathcal{S}$, is described by the sequence of board arrangements ($B_i$) and associated actions ($A_i$) leading to the current board, $(B_0,A_0,\dots,B_t)$. The game engine generator, $g$, randomly generates the initial board arrangement $B_0$ per input parameters provided for the experiment, $\beta$, i.e., $S_0=(B_0)\sim g(\beta)$. The action space $\mathcal{A}$ is the set of 144 action tuples $(r,c,b)$ given by placing the piece in row $r\in\{1,\dots,6\}$ and column $c\in\{1,\dots,6\}$ into bucket $b\in \{0,\dots,3\}$. If the game engine evaluates that action $A_t$ is permitted by the rule, the corresponding piece is removed in board arrangement $B_{t+1}$, otherwise $B_{t+1}=B_t$. Note that state transitions are deterministic given the player's action, according to the logic of the hidden rule. A player receives reward $R_t(S_t,A_t)=0$ if action $A_t\in\mathcal{A}$ from state $S_t$ is permitted by the hidden rule and $-1$ otherwise. A terminal state is reached when the board is cleared, denoted by $t=T$. The player's objective is to find a policy $\pi(s)$ that maximizes the value function $v_{\pi}(s)=\mathbb{E}_{\pi}[\sum_{k=t}^T\gamma^{k-t}R_k(s,\pi(s))|S_t=s]$ over all $s\in \mathcal{S}$ reachable from $S_0\sim g(\beta)$, where $\gamma \in (0,1)$ is a discount factor.

We used two sample RL algorithms for comparison to human players. As an example of a policy-gradient based method, we employed a variant of the canonical REINFORCE algorithm \citep{sutton_simple_1992}. For a sample value-based method, we used a variant of epsilon-greedy DQN with experience replay \citep{mnih_human-level_2015}. $\mathcal{S}$ is too large to deal with directly and thus we constructed feature maps $\phi_i(s)$ to make the problem tractable. Our goal in constructing $\phi_i$ was to faithfully represent $\mathcal{S}$ without artificially biasing algorithm performance for any particular subset of rules. As part of our experiments, we explored numerous feature maps and found that none universally outperformed the others across our tested rules (see Section~\ref{sec:results}). For REINFORCE, we used a neural network, parameterized by $\theta$, to represent the learner's policy. Similarly for DQN, we used such a neural network to approximate the state-action value function $Q(S,A)$. Complete descriptions of these algorithms, corresponding hyperparameter selections, feature maps, and experimental flow can be found in Appendix~\ref{app:algos}. For a fair comparison to humans encountering the GOHR for the first time, we measured the performance of these RL algorithms with no pre-training. Each \textbf{learning run} for a given algorithm and rule consisted of initialization of the model with random network weights followed by serial play of a set number of episodes of the same rule. Learning runs used separate random seeds for the algorithm and the game-engine board generator. For each algorithm, we performed 50 independent learning runs on each rule provided in Section~\ref{sec:conceptual_framework}.

\textbf{Performance metrics and statistical comparison }
Rules may permit many sufficient policies; we measured performance based on a learner's ability to exhibit \textit{any} sufficient policy for the rule. Due to the limited participation time of human players, we measured a human player's understanding of a rule using streaks of consecutive correct moves that are sufficiently unlikely to occur at random. The base rules give players a $\nicefrac{1}{4}$ chance of making a correct move at random; we chose a threshold of 10 correct moves in a row as it corresponds to a random probability of roughly $1\times10^{-6}$. We define the point metric $m^*$ to be the index of the first move in the first streak of 10 or more correct moves that a player demonstrates. If a player never achieves such a streak, they are assigned an arbitrary placeholder value for $m^*$ that is higher than all measured $m^*$ values across the population of human players. Larger values of $m^*$ are interpreted to mean greater difficulty as they correspond to the player needing more moves to obtain an understanding of the rule.

\begin{figure}[htbp]
    \begin{minipage}{0.4\textwidth}
        \includegraphics[width=.8\textwidth]{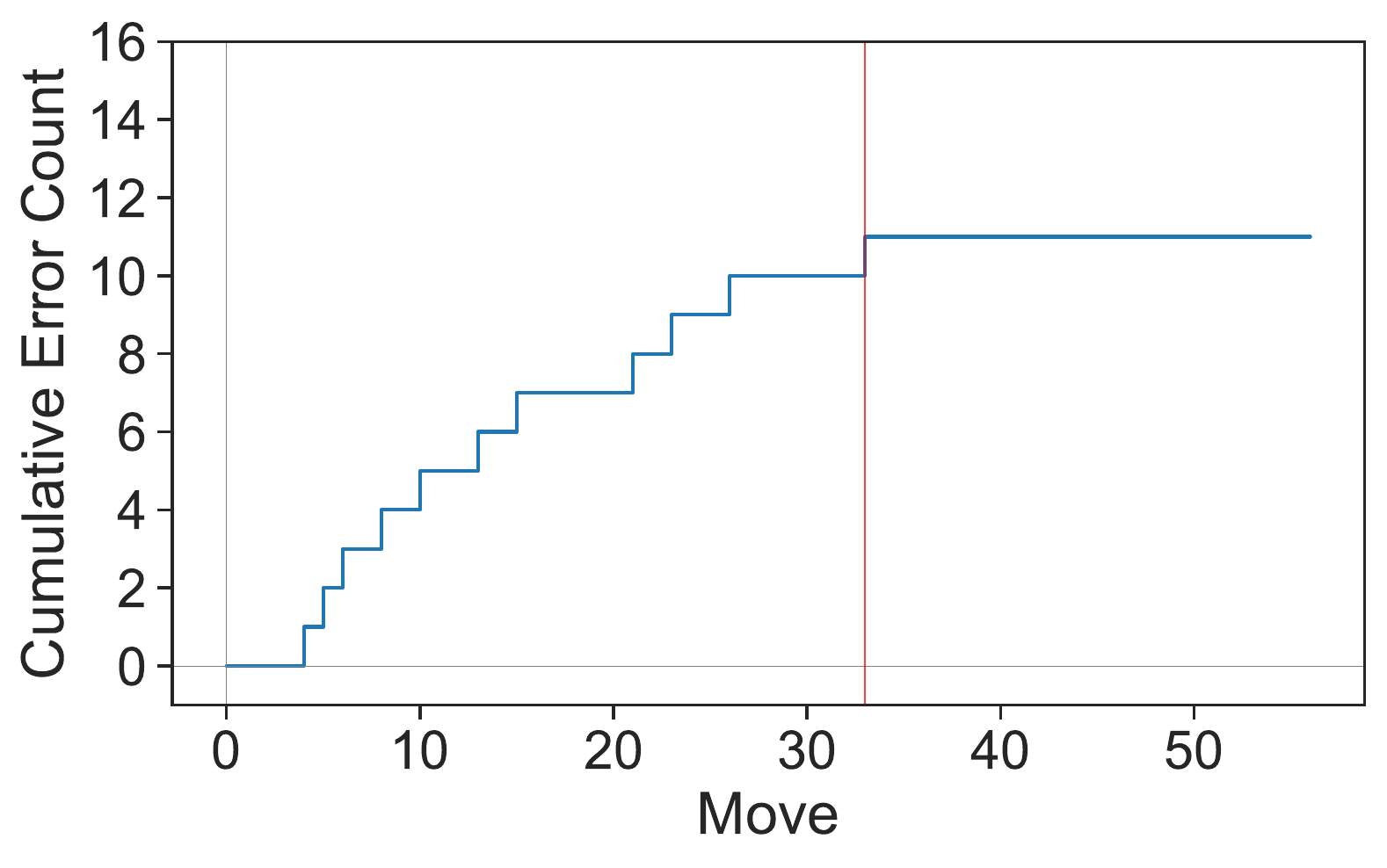}
    \end{minipage}
    \hspace{.25cm}
    \begin{minipage}{0.4\textwidth}
        \includegraphics[width=.8\textwidth]{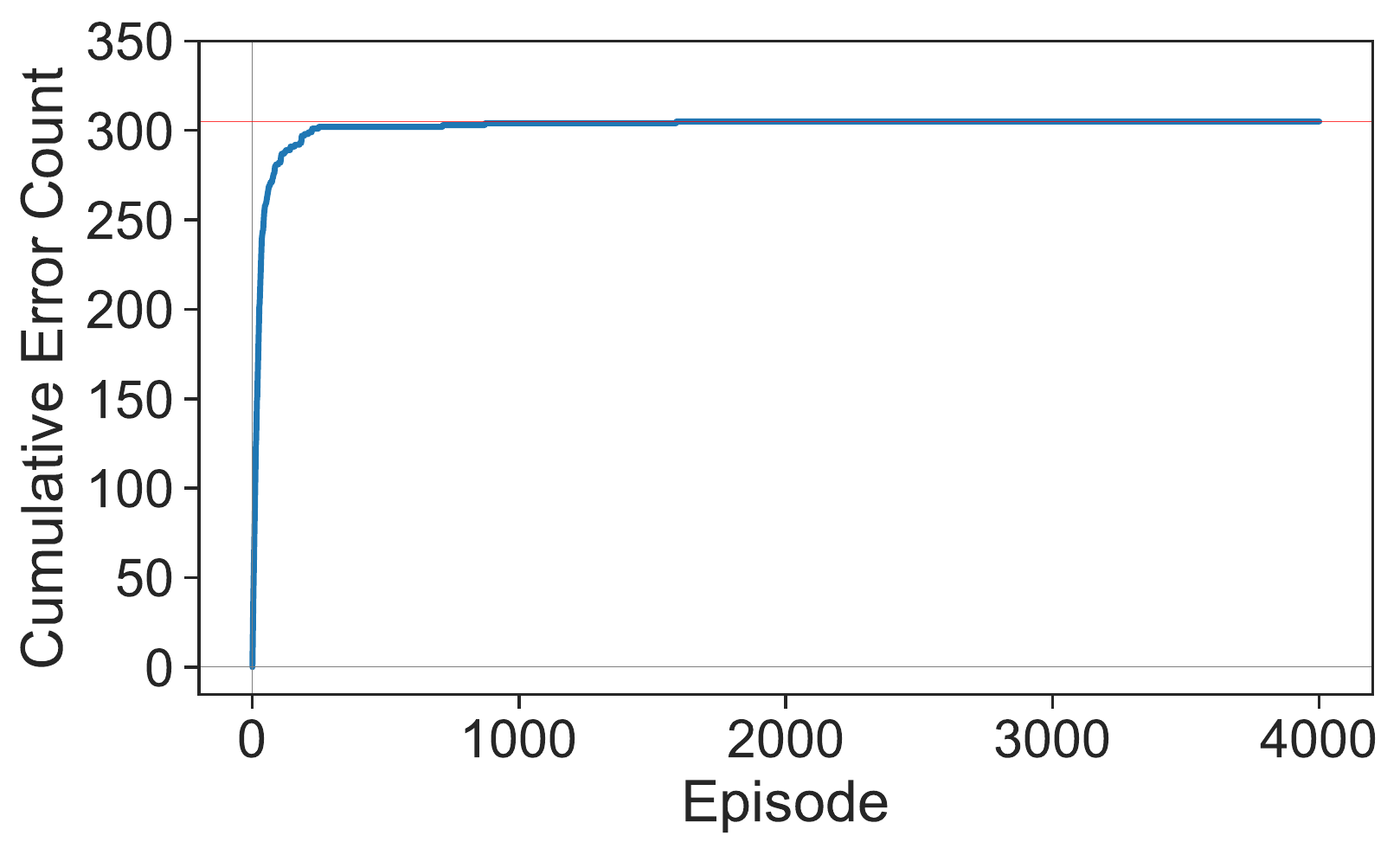}
    \end{minipage}
    \centering
    \caption{Sample learning runs for a human (left) and DQN (right), plotting cumulative error count against the move and episode indices, respectively. The human's learning run is summarized by an $m^*$ of 34, the first move of the first streak of 10+ correct moves. The DQN's learning run is summarized by a TCE of 305, the error count after 4000 episodes.}
    \label{fig:sample_player}
\end{figure}

In contrast to humans, we can prompt RL algorithms to play a fixed number of episodes large enough to exhaustively evaluate policy sufficiency. We define the point metric of terminal cumulative error (TCE) to be the cumulative error count made across all episodes in a learning run. If the algorithm has reached an understanding of the hidden rule, this error count is expected to converge to a constant, with some allowance made for algorithm stochasticity (see Appendix~\ref{app:algos_metrics} for chosen convergence criteria). If a learning run does not meet our convergence criteria, we set the TCE for that learning run to be a placeholder value larger than all convergent TCE values. We chose fixed episode horizons of 4,000 and 60,000 for learning runs of DQN and REINFORCE, respectively; learning runs typically converged well before these horizons, but use of a common horizon allowed for fair comparison of learning runs that needed more episodes to converge. As with $m^*$, larger values of TCE are interpreted as the result of more difficult learning tasks. The metrics $m^*$ and TCE summarize the performance of each learning run, as shown in Figure~\ref{fig:sample_player}. For each type of player (humans, DQN, and REINFORCE), we gather their associated $m^*$ or TCE values for all learning runs on all rules and we compare their distributions using the non-parametric Mann-Whitney U-Test \citep{nachar_mann-whitney_2008}. Statistical tests are performed at the $\alpha = 0.05$ significance level.

\section{Results}\label{sec:results}
\textbf{Comparison of feature representations } As noted, our RL experiments considered different feature representations of the observed state. To study the impact of memory on performance, we tested input feature maps that included 2, 4, 6, or 8 previous board states and actions. This testing further included feature maps with different representations of both the past boards and actions themselves. For example, a past action could be represented as a 144-long one-hot vector or by three one-hot vectors (6-long for row, 6-long for column, and 4-long for bucket). We refer to the former as a sparse representation and the latter as a dense representation; we extend similar notions to representations of past boards. We found that no feature map universally outperformed the others across all tested rules; different choices of memory and board/action representation yielded different performance tradeoffs. In general, additional memory improved performance on non-stationary rules but worsened performance on stationary rules. With some exceptions, dense representations tended to outperform sparse representations. Additionally, while REINFORCE performed poorly in comparison to DQN, we noted that both algorithms showed similar trends in performance with respect to different feature representations. We provide a complete discussion of performance using different feature maps in Appendix~\ref{app:feat_discussion}. In order to provide a single point of comparison for each method to humans in the following experiments, we select a feature map that showed a good balance of performance across our tested rules for both algorithms: dense board and action representations with 6 steps of memory.

\begin{figure}[htbp]
    \begin{minipage}{0.45\textwidth}
        \centering
        \includegraphics[width=.9\textwidth]{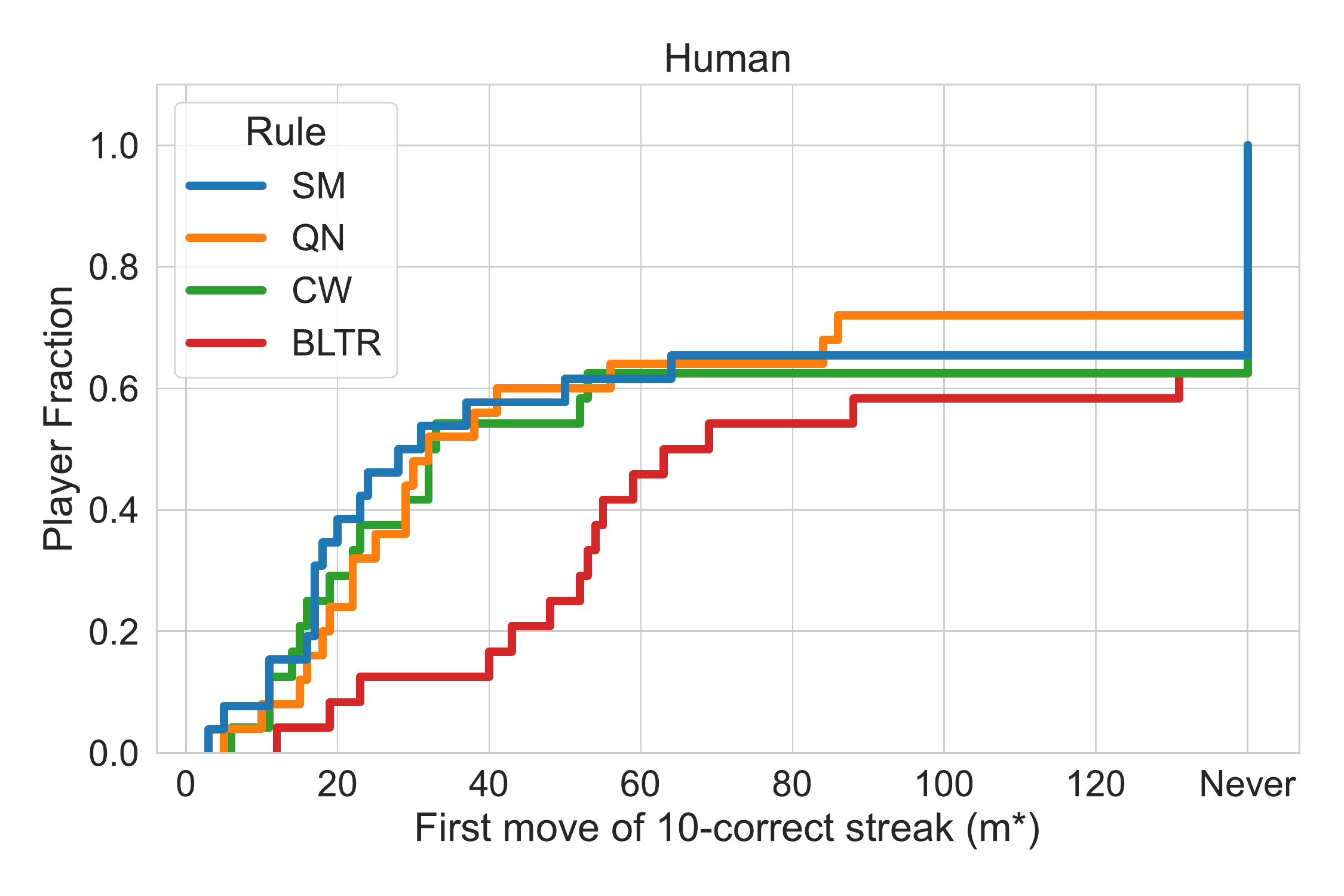}
    \end{minipage}
    \begin{minipage}{0.5\textwidth}
        \centering
        \includegraphics[width=.7\textwidth]{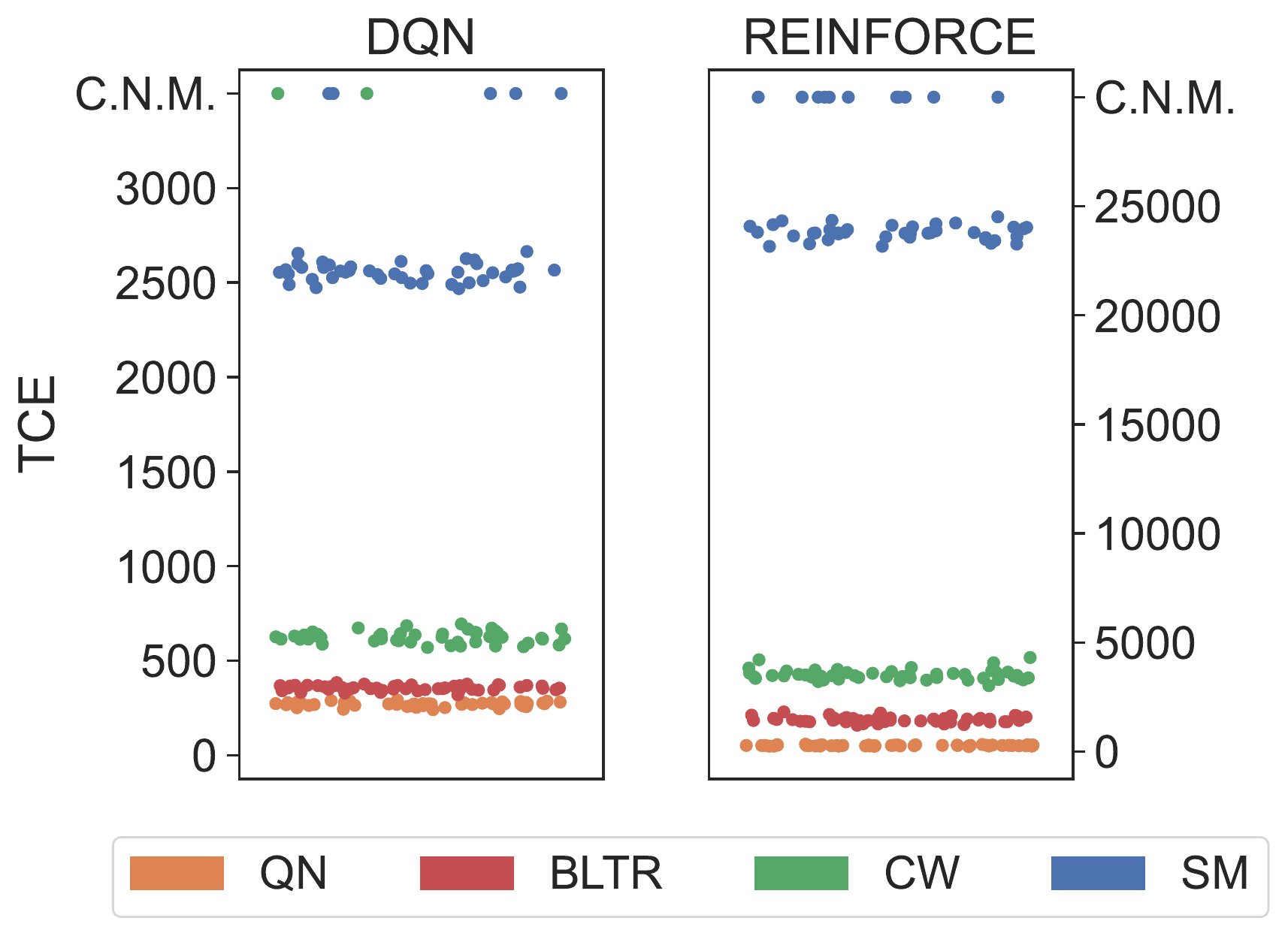}
    \end{minipage}
    \caption{Base rule performance of humans (left) and RL players (right). ECDF curves denote the fraction of human players achieving an $m^*$ streak by a given move index on each rule (`Never' indicates player does not achieve such a streak). Strip plots of TCE distributions of each rule are provided for DQN and REINFORCE, separated due to different TCE magnitudes (`C.N.M.' indicates convergence criteria were not met for that learning run). Each dot corresponds to a learning run.}
    \label{fig:base-performance}
\end{figure}

\textbf{Comparison of base rules }
Our first experiment introduced a set of `base rules' for study (SM, QN, CW, BLTR). Figure~\ref{fig:base-performance} summarizes performance on these rules using empirical cumulative distribution function (ECDF) plots of $m^*$ for humans and strip plots of TCE for DQN and REINFORCE. Note that lower ECDF curves indicate greater difficulty as they imply that fewer players achieve an $m^*$ streak at a given move index. Higher values of TCE indicate greater difficulty as more mistakes occur before reaching a sufficient policy. See Appendix~\ref{app:base_rule_compare} for a complete tabulation of p-value results from two-sided Mann-Whitney U-tests for all base rule pairs and learners.
\\First, we note the similarity of human performance across the base rules, despite differences in learning task structure. Only one rule pairing, QN-BLTR, showed a statistically significant difference in human performance, with BLTR appearing more difficult. In contrast, both DQN and REINFORCE showed statistically significant performance differences for all rule pairs, even after applying a conservative Bonferroni correction. Although DQN outperformed REINFORCE (measured by TCE), both algorithms exhibited the same rule difficulty ordering: QN, BLTR, CW, SM (easiest to hardest, with SM noticeably more difficult). The non-stationary rules followed an expected ordering based on underlying sequence length: BLTR, a 2-long sequence, was easier than CW, a 4-long sequence. Regarding stationary rules, we did not expect SM to be so difficult, especially compared to QN. We believe the difficulty of SM reflects a subtle interaction between our feature representations and the details of the learning task. While our boolean board representations are an intuitive way to create a static size characterization of the board, they favor the learning of position-based patterns over shape- or color-based patterns. In particular, the one-hot encoding fails to provide a notion of similarity for identical shapes or colors in different board positions. This type of finding might be easily overlooked in settings outside the GOHR, where the relevant learning tasks are more opaque and systematic investigation of task structure is not the primary focus.

These results suggest meaningful differences between HL and RL in their response to varying task structures. In particular, human performance likely depends on both the structure of the task and its relation to human priors for patterns. The plausible closeness of these rules to common priors (e.g. clockwise) would explain the similar human performance we observed across rules despite their structural differences. RL players, on the other hand, responded strongly to differences in the logical structure of learning tasks and performed identically for logically equivalent rules, such as SM and its color-based equivalent CM (see Appendix~\ref{app:color_rules}). The GOHR serves as an ideal testbed for deeper investigations into such differences in task structure. For example, CW represents one possible instance of a 4-long repeating pattern; a dedicated experiment might explore other 4-long patterns to measure how human priors affect performance across rules of equivalent logical structure. Likewise, such an experiment could also include 2-, 3-, or 5-long patterns to precisely measure how human and RL player performance varies with respect to incremental changes in the logical structure itself. Similar approaches could be used to explore performance differences within families of stationary rules and, more generally, to identify the strengths and weaknesses of both learners with respect to fundamental elements of task structure. Further, even from this example experiment we see that the difficulty ordering of a common set of rules is not shared for human and RL players, suggesting that human-machine learning pairs might be constructed to exploit the differing strengths of each.

\begin{figure}[t]
    \begin{minipage}{0.43\textwidth}
        \includegraphics[width=.9\textwidth]{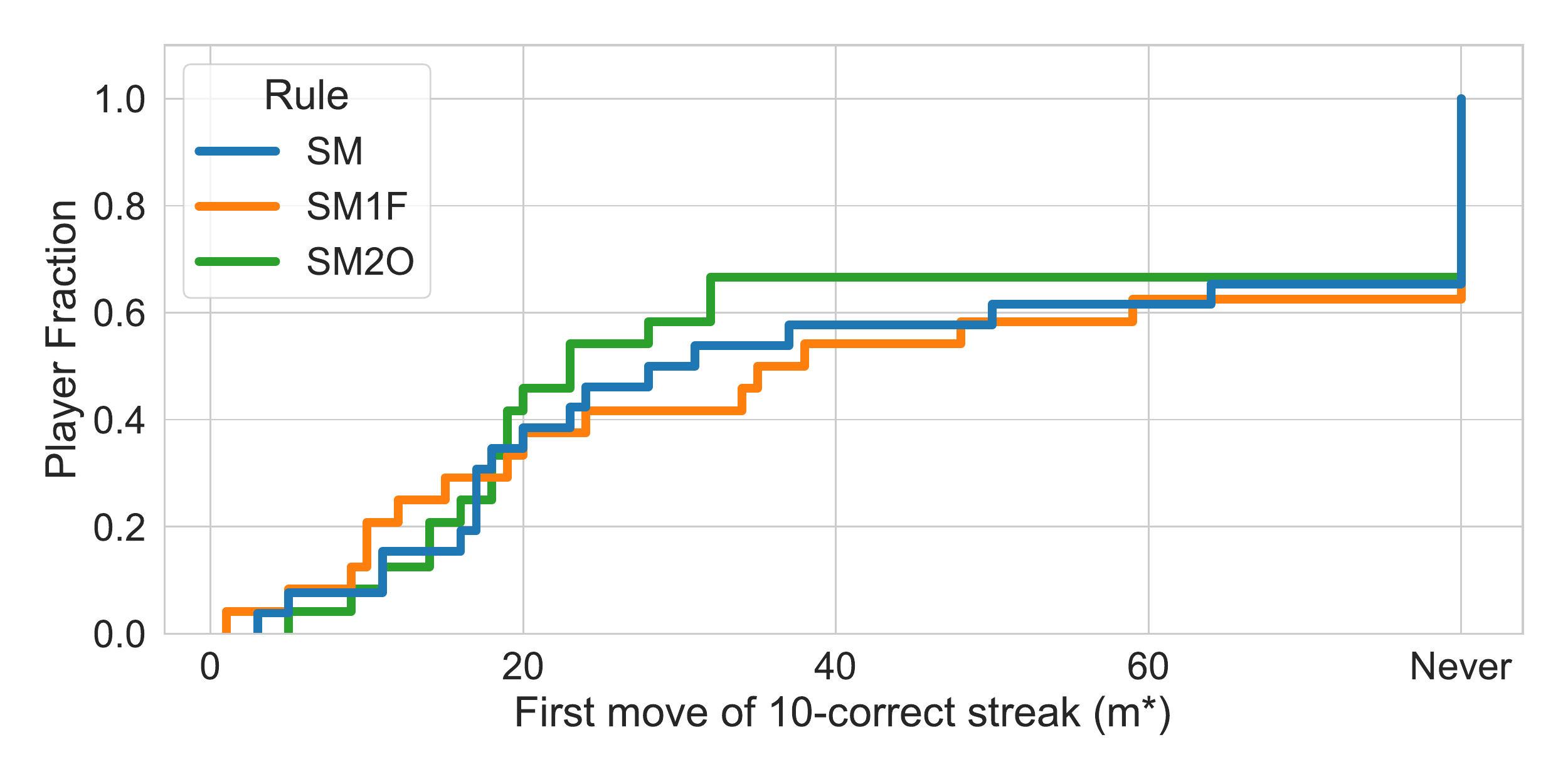}
    \end{minipage}
    \hspace{\fill}
    \begin{minipage}{0.43\textwidth}
        \includegraphics[width=.9\textwidth]{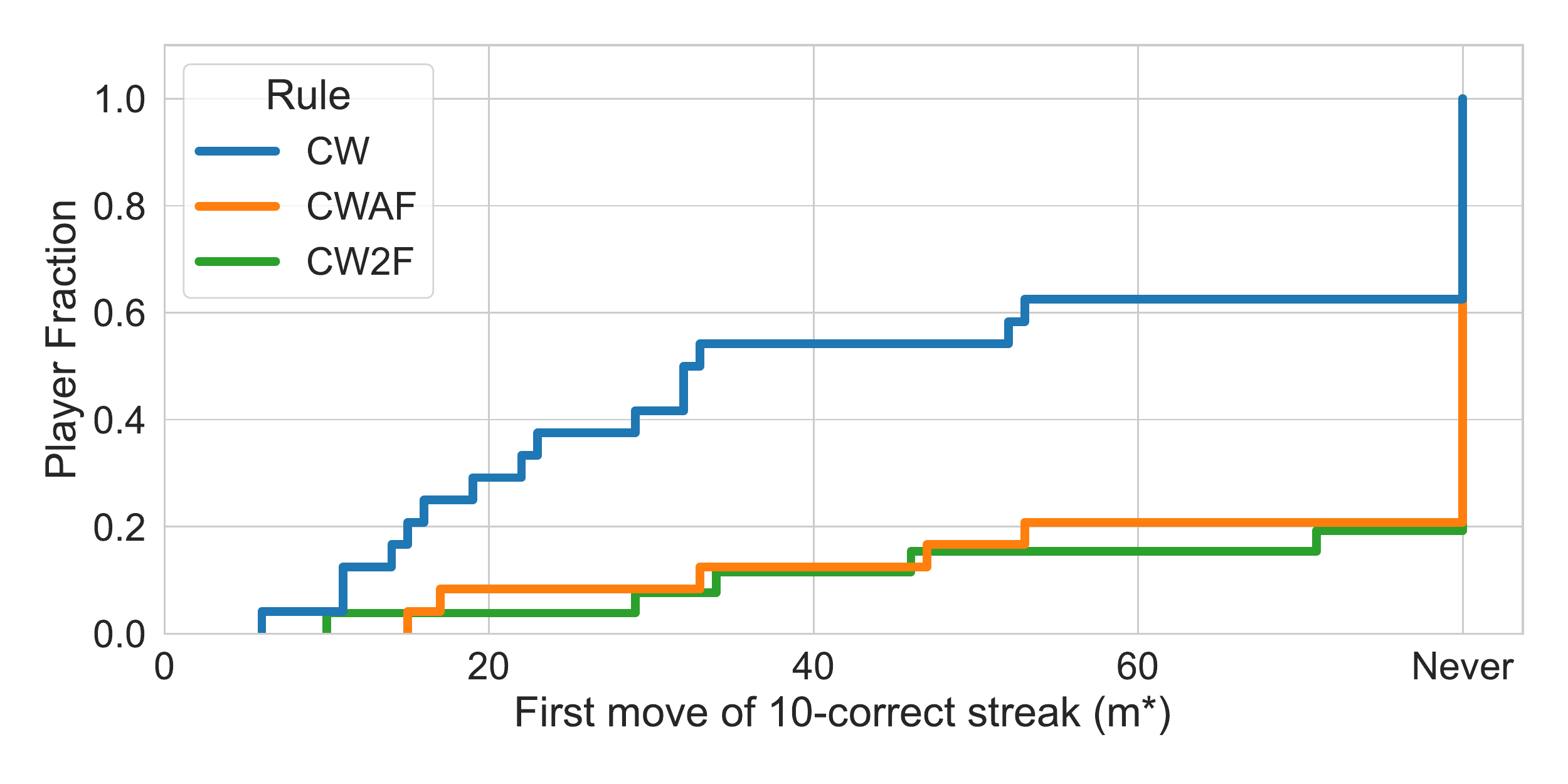}
    \end{minipage}
    \begin{minipage}{0.43\textwidth}
        \includegraphics[width=.9\textwidth]{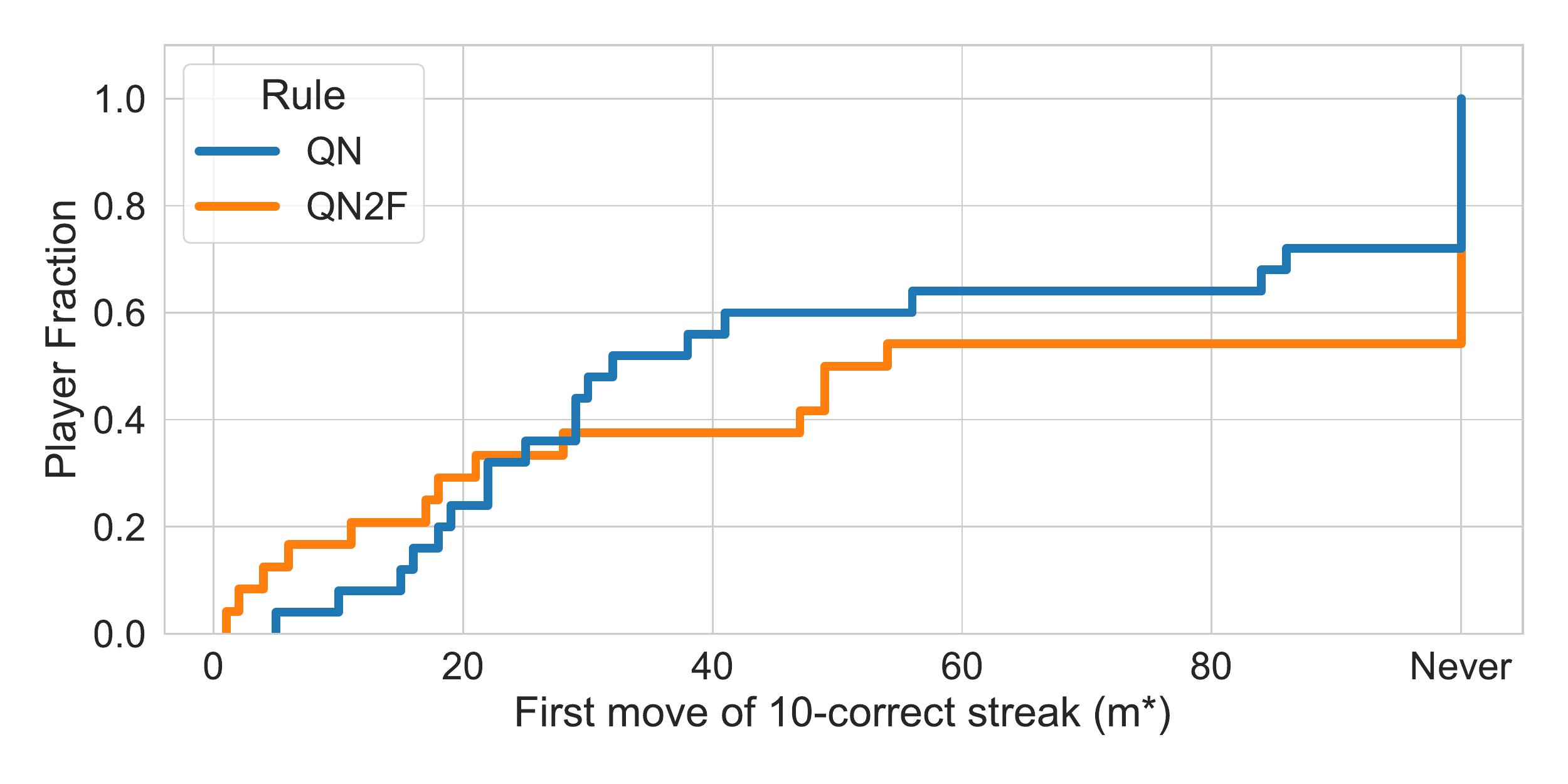}
    \end{minipage}
    \hspace{\fill}
    \begin{minipage}{0.43\textwidth}
        \includegraphics[width=.9\textwidth]{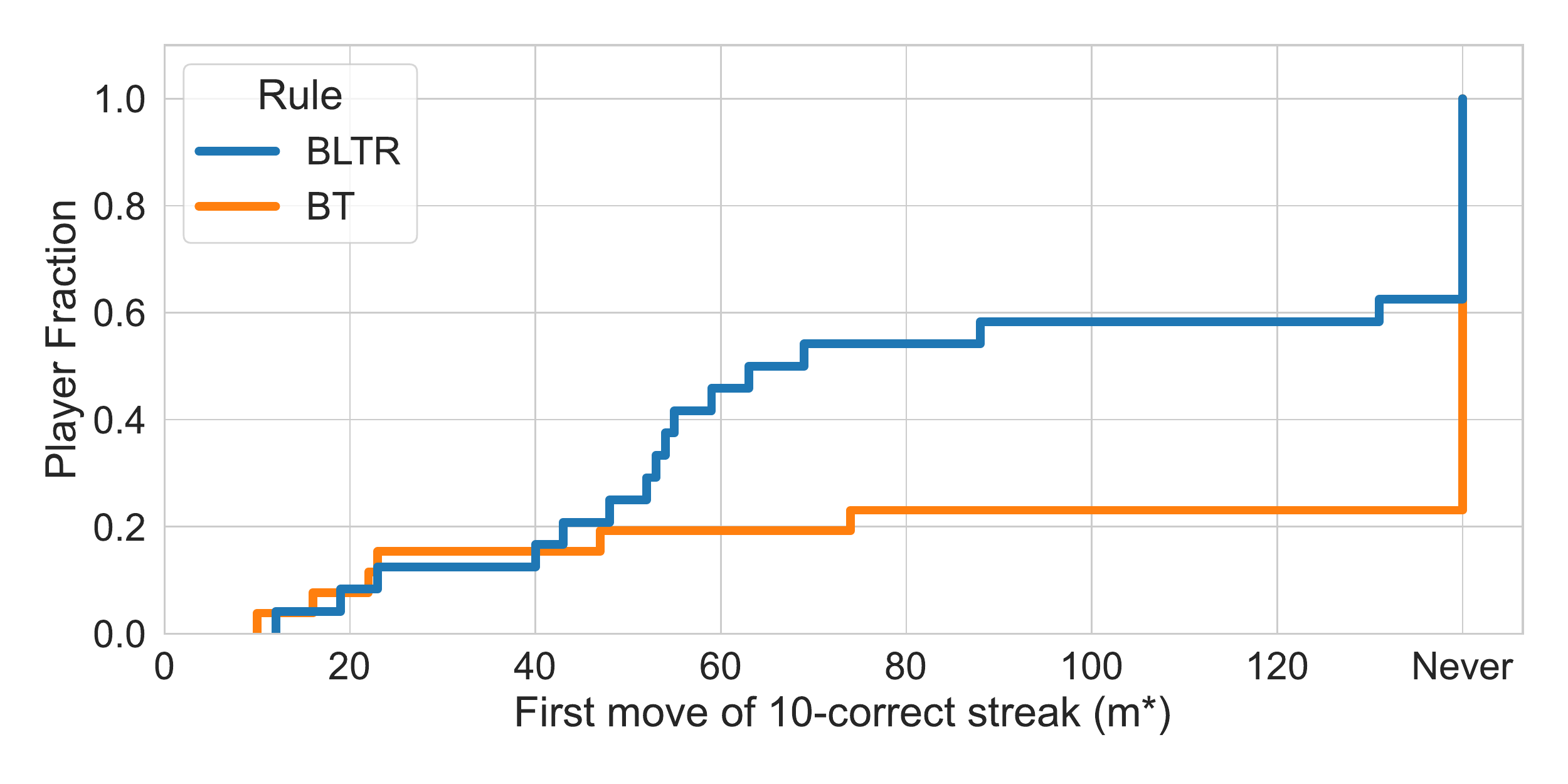}
    \end{minipage}
    \caption{ECDFs of $m^*$ distributions for each rule family. Stationary rules (shape/quadrant) shown on left. Non-stationary rules (clockwise/alternating) shown on right. Base rules shown in blue.}
    \label{fig:human_ecdfs}
\end{figure}

\textbf{Impact of rule generality }
Our second experiment introduced more general variants of our base rules. Families with three rules (e.g., base rule $\mathcal{A}$ and more general rule variants $\mathcal{B},\mathcal{C}$) offer two generality comparisons ($\mathcal{B}\succ \mathcal{A}$ and $\mathcal{C}\succ\mathcal{A}$), while families with two rules offer one comparison ($\mathcal{B}\succ \mathcal{A}$). Figure~\ref{fig:human_ecdfs} summarizes human performance within each rule family with ECDFs of the $m^*$ distributions. DQN and REINFORCE showed uniformly better performance on more general rule variants (see Appendix~\ref{app:rl_rule_gen} for related plots). Table~\ref{tab:generality-stat-tests} shows the results of the Mann-Whitney U-Tests associated with each generality comparison. We used one-sided tests, with the null hypothesis that the more general rule is no harder than the base rule, as more general rules offer a larger number of sufficient policies. Per the U-tests, we note that RL players uniformly found more general rule variants easier than their base counterparts (the p-values greater than 0.999 would be significant under the opposite direction null hypothesis). In contrast, the response of human players to increasing generality depended on the structure of the base rule. In particular, human players appeared to find the more general forms of our non-stationary rules more difficult than their base rule counterparts (i.e., CWAF and CW2F appear more difficult than CW, BT appears more difficult than BLTR). This is surprising as more general rules offer a higher probability of achieving a streak of 10 correct moves at random. We posit that this difference between humans and our RL players reflects important differences in their respective learning strategies. While greater rule generality might plausibly assist learning by offering a larger number of sufficient policies for the player to learn, it also could hinder learning by decreasing the amount of useful, negative feedback. For primarily inductive learners, such as our sample RL algorithms, it appears that the availability of a larger number of sufficient policies dominates, making these more general rules easier. Humans, however, likely employ some combination of induction and deduction; the additional positive feedback from more general rules may complicate deduction as feedback could agree with many candidate classes of hidden rules. Future studies, with a broader set of base rules, could explore such an effect in greater detail. As in the first experiment, HL and RL did not respond identically to changes in task structure, and our results show that the parallel study of task structure for HL and RL may provide important insight into the strengths and weaknesses of each learner. Although the GOHR deals with relatively abstract task structures, we believe a systematic understanding of performance within the GOHR can provide important perspective in complex environments, where tasks are compositions of many such fundamental elements.

\begin{table}[h]
\centering
\caption{P-value results of one-sided U-Tests comparing each more general rule to its base rule counterpart. Tests use $m^*$ and TCE as point metrics for humans and algorithms, respectively. Null hypothesis is always that the more general rule is no harder with alternative hypothesis that the more general rule is harder. Significant results at the $\alpha=0.05$ level are highlighted with a $\dag$ and columns with contrasting HL/RL behavior are shown in bold.}
\label{tab:generality-stat-tests}
\begin{tabular}{@{}lcccccc@{}} \toprule
& SM1F & SM2O  & QN2F & \textbf{BT} & \textbf{CWAF} & \textbf{CW2F}\\ 
& \multicolumn{2}{c}{$(\succ$ SM)} & $(\succ$ QN) & \textbf{$(\succ$ BLTR)} & \multicolumn{2}{c}{\textbf{$(\succ$ CW)}}\\\midrule
Human & 0.496 & 0.620 &  0.322 & 0.014$^\dag$ & 0.001$^\dag$ & <0.001$^\dag$ \\
DQN  & 0.999 & 0.999 &  0.999 & 0.999 & 0.999 & 0.999 \\
REINFORCE & 0.999 & 0.999 &  0.999 & 0.999 & 0.999 & 0.999 \\
\bottomrule
\end{tabular}
\end{table}

\section{Conclusion}\label{sec:conclusion}
We have shown that the GOHR provides a capability for studying the performance of HL and RL in a novel and principled way. Using the GOHR's expressive rule syntax, researchers can make precise changes to learning tasks in order to study their effects on human and RL algorithm performance. The GOHR complements existing environments by empowering researchers to perform rigorous experiments into different learning task structures. Beyond the kind of experiments presented here, the GOHR could also be used for related studies such as teaching curricula, transfer learning, or human-machine learning pairs. Task-oriented experiments augment efforts to improve the overall capabilities of RL algorithms by furthering our understanding of these methods' strengths and weaknesses. Most importantly, this type of study provides a step toward task-oriented understandings of RL \emph{and} HL, both of which are needed to better inform the real-world use of RL. With this goal in mind, we are sharing the complete suite of tools with all interested researchers. We hope that researchers using the GOHR will share their findings to help this inquiry.

\begin{ack}
Support for this research was provided by the University of Wisconsin-Madison Office of the Vice Chancellor for Research and Graduate Education with funding from the Wisconsin Alumni Research Foundation, and by the National Science Foundation under Grant No. 2041428. Any opinions, findings, and conclusions or recommendations expressed in this material are those of the author(s) and do not necessarily reflect the views of the sponsors. We acknowledge numerous stimulating discussions with collaborators in the broader project: Professors Xiaojin (Jerry) Zhu and Gary Lupyan; Drs. Ellise Suffill and Charles Davis; and doctoral student Yuguang (Aria) Duan. We thank Mr. Kevin Mui for building the first instance of the GOHR, to support research on human learning. We also thank doctoral students Shubham Bharti and Yiding Chen for their work on past machine learning experiments and associated code.
\end{ack}

\bibliographystyle{plainnat}
\bibliography{gohr_2023}


\newpage
\appendix

\part{Appendix}
\addcontentsline{toc}{section}{Appendix}
\parttoc
\newpage
\section{Appendix}

\subsection{Preliminary work}\label{app:prev_work}
Preliminary work on the GOHR \citep{pulick_game_2022} introduced the environment and preliminary machine learning experiments. As noted in the body of the text, some sections of this manuscript (particularly descriptions of the environment and machine learning performance metrics) follow closely from corresponding sections in \citep{pulick_game_2022}. This manuscript presents a broader set of experiments, includes new reinforcement learning models, adds featurization studies, and introduces comparisons to human learners in order to demonstrate the full capability of the GOHR as a research tool. 

\subsection{GOHR Mechanics}\label{app:gohr}
In this section we provide additional details, as discussed in the body of the manuscript, regarding the structure and mechanics of the GOHR. Additional details can be found in the external documentation at our \href{http://sapir.psych.wisc.edu:7150/w2020/captive.html}{public site}.

\subsubsection{Board generation}\label{app:gohr_boards}
The board generation process (whether using pre-defined boards or done randomly) is controlled through the configuration file defined for each experiment. If using pre-defined boards, the experiment designer must specify a collection of JSON board representations as part of this file. When generating boards randomly, the experimenter must specify the minimum and maximum numbers of game pieces, colors, and shapes to appear on each new board; the game engine randomly selects values in the range [minimum, maximum] for each quantity and generates boards accordingly. 

\subsubsection{Rule specification and examples}\label{app:gohr_rules}
This overview of GOHR rule specification closely follows \cite{pulick_game_2022}. In Section~\ref{sec:gohr}, we considered the example where stars and triangles always go in the top-left (0) bucket while circles and squares always go in the bottom-right bucket (2), regardless of their color or position. As noted, this can be expressed with two atoms on a single line:
\begin{center}
    (*, [star,triangle], *, *, 0) (*, [circle,square], *, *, 2)
\end{center}
A more complex example, where stars and triangles go in the top-left bucket (0), squares in the bottom three rows go in the top-right bucket (1), squares in the top three rows go in the bottom-right bucket (2), and circles go in either of the bottom buckets (2,3), could be constructed with the following four atoms on a single line:
\begin{center}
    (*,[star,triangle],*,*,0) (*,square,*,[R1,R2,R3],1) (*,square,*,[R4,R5,R6],2) (*,circle,*,*,[2,3])
\end{center}
For brevity, we use shorthand to refer to each row of the game board (R\#) rather than individual board cells. Using this approach, an experimenter can design stationary hidden rules which are complex combinations of game piece features (shape, color, or position).

As mentioned, each atom contains a count field dictating the number of corresponding correct moves that the atom permits. When a player makes a move satisfying one (or more) of the atoms on the active rule line, the counts associated with the satisfied atoms are decremented by one. An atom with a count of 0 is considered exhausted and no longer accepts moves. At each move, the game engine evaluates the active rule line and the current board; if there are no valid moves available to the player, the engine makes the next rule line active and resets all counts in the new line. If there are no lines below the current rule line, the first rule line is made active again. This functionality can be used to encode sequences into the hidden rules. For example, the following two rules require the player to alternate between placing game pieces in the top (0,1) and bottom (2,3) buckets, with subtly different mechanics.

\begin{center}
    \begin{tabular}{c c}
        Strict Alternation & Ambiguous Alternation \\ \hline
	(1, *, *, *, [0,1]) & (1, *, *, *, [0,1]) (1, *, *, *, [2,3])  \\
        (1, *, *, *, [2,3]) &  
    \end{tabular}
\end{center}

In the strict case, the player's first move is allowed into only the top buckets and every correct move exhausts the current rule line. As a result, the active rule line will alternate with each correct move until the board is cleared. In the ambiguous case, a similar alternation occurs, but the order depends on the player's first correct move. Since both atoms are on the same rule line, the player's first move may go in any of the four buckets. After one successful move, the player may only make a move satisfying the remaining, non-exhausted atom. After two successful moves, all atoms in the rule line are exhausted and both are reset. This process repeats until the board is cleared.

GOHR also permits the experimenter to attach a count to each rule line. When the line is active, this count decrements each time any atom in the line is satisfied. For instance, a rule that alternates between uniquely assigning shapes and colors to buckets would look like:
\begin{center}
    1 (*, star, *, *, 0) (*, square, *, *, 1) (*, circle, *, *, 2) (*, triangle, *, *, 3)\\
    1 (*, *, black, *, 0) (*, *, yellow, *, 1) (*, *, blue, *, 2) (*, *, red, *, 3)
\end{center}
If the rule-line count is exhausted, the game engine makes the next line active, even if there are non-exhausted atoms on that line. For this example, the active rule line will alternate after each successful move, regardless of which atom in the active line is satisfied. If no count is provided for a given rule line, the game engine assumes that the rule line is not metered and can be used until all atoms on that line are exhausted or no valid move exists for the game pieces currently on the board.

The GOHR rule syntax allows the experimenter to write expressions in an atom's bucket field that define buckets based on previous successful moves. The game engine stores values for the bucket that most recently accepted any game piece ($p$) as well as the bucket that most recently accepted an item of each color ($pc$) and shape ($ps$). A simple rule expressible using these values is ``objects must be placed in buckets in a clockwise pattern, beginning with any bucket'':

\begin{center}
    (1, *, *, *, [0,1,2,3]) \\
    (*, *, *, *, p+1)
\end{center}

The expressions used in the buckets field are evaluated modulo 4 to ensure that the resultant expression gives values in the range 0-3. The game engine also supports the terms ``nearby'' and ``remotest'' as bucket definitions, which allow the experimenter to require that game pieces be put into either the closest or furthest bucket, respectively, evaluated by Euclidean distance. 

The arrangement of atoms allows the experimenter to encode a specified order of the component tasks within a rule. For instance, the rule that ``all red pieces go into bucket 1, \emph{then} all blue pieces go into bucket 2'' would be expressed as follows:
\begin{center}
    (*, *, red, *, 1) \\
    (*, *, blue, *, 2)
\end{center}
Since both the first rule line and associated atom are not metered, they can never become exhausted, even if there are no more red game pieces left on the board. However, as noted above, the engine evaluates if there are any valid moves available to the player given the active rule line; if there are no valid moves available, the engine makes the next line active. In this example, once the player has cleared all red game pieces from the board, the engine will make the second rule line active.

\subsection{Sample experiment rules in GOHR syntax}\label{app:rule_syntax}
In this section we provide the GOHR rule syntax associated with each of the rules included in our experiments. Note that the default set of four colors is red, blue, yellow, and black. Our human experiments substituted black for green to give more visual contrast between different colors. The syntax below is provided for the human experiments; substituting black for green (where applicable) gives the syntax for the versions played by RL players. Since our RL players do not process a visual representation of the board, the specific choice of the four colors used for RL play has no impact on performance.

\textbf{Shape Match (SM)}:
\begin{center}
    (*, star, *, *, 0) (*, triangle, *, *, 1) (*, square, *, *, 2) (*, circle, *, *, 3)
\end{center}

\textbf{Shape Match 1 Free (SM1F)}:
\begin{center}
    (*, star, *, *, [0,1,2,3]) (*, triangle, *, *, 1) (*, square, *, *, 2) (*, circle, *, *, 3)
\end{center}

\textbf{Shape Match 2 Options (SM2O)}:
\begin{center}
    (*, star, *, *, [1,3]) (*, triangle, *, *, [0,2]) (*, square, *, *, [1,3]) (*, circle, *, *, [0,2])
\end{center}
Note that SM2O could structurally be formed using many different 2-option bucket arrangements. This particular bucket arrangement technically does not properly dominate SM, although it does dominate a version of SM with a slightly different bucket arrangement. This disconnect was due to the fact that some experiments not discussed in this paper considered the performance of players when playing episodes of these rules in particular orders. For players who might see both SM and SM2O as part of an experiment, we wanted to ensure that players could not simply carry over their policy from a prior rule to a new rule and obtain error-free play (as would be the case if playing SM and SM2O sequentially for a version of SM2O that dominates SM). Given no compelling reason for a different arrangement of buckets to impact performance for human players or RL players, we therefore still consider SM2O to be a more general version of SM for the purposes of the sample experiments included in this paper.

\textbf{Color Match (CM)}:
\begin{center}
    (*, *, green, *, 0) (*, *, yellow, *, 1) (*, *, red, *, 2) (*, *, blue, *, 3)
\end{center}

\textbf{Color Match 1 Free (CM1F)}:
\begin{center}
    (*, *, green, *, 0) (*, *, yellow, *, 1) (*, *, red, *, [0,1,2,3]) (*, *, blue, *, 3)
\end{center}

\textbf{Color Match 2 Options (CM2O)}:
\begin{center}
    (*, *, green, *, [1,3]) (*, *, yellow, *, [0,2]) (*, *, red, *, [1,3]) (*, *, blue, *, [0,2])
\end{center}
Note the same point for CM2O and CM as discussed for SM2O and SM. We see no compelling reason for the specific choice of buckets to impact performance and thus still consider CM2O to be a more general version of CM.

\textbf{Quadrant Nearby (QN)}:
\begin{center}
    (*,*,*,*,Nearby)
\end{center}

\textbf{Quadrant Nearby 2 Free (QN2F)}:
\begin{center}
    (*,*,*,[19,20,21,22,23,24,25,26,27,28,29,30,31,32,33,34,35,36],Nearby)\; \textbackslash\\(*,*,*,[1,2,3,4,5,6,7,8,9,10,11,12,13,14,15,16,17,18],[0,1,2,3])
\end{center}
Note that the \textbackslash\; denotes that these two atoms are on the same rule line and are only separated here due to formatting constraints.

\textbf{Bottom-left then Top-Right (BLTR)}:
\begin{center}
    (1,*,*,*,[3])\\
    (1,*,*,*,[1])
\end{center}

\textbf{Bottom then Top (BT)}:
\begin{center}
    (1,*,*,*,[2,3])\\
    (1,*,*,*,[0,1])
\end{center}

\textbf{Clockwise (CW)}:
\begin{center}
    (1,*,*,*, 0)\\
    (*,*,*,*, p+1)
\end{center}

\textbf{Clockwise Alternating Free (CWAF)}:
\begin{center}
    (1,*,*,*,0)\\
    (1,*,*,*,[0,1,2,3])\\
    (1,*,*,*,2)\\
    (1,*,*,*,[0,1,2,3])
\end{center}

\textbf{Clockwise 2 Free (CW2F)}:
\begin{center}
    (1,*,*,*, 0)\\
    (1,*,*,*, 1)\\
    (2,*,*,*, [0,1,2,3])
\end{center}

\subsection{Human experimental procedures}\label{app:human_exp}
In this section we provide additional information regarding our experiments with human participants.

\subsubsection{Overview}
Our Amazon Mechanical Turk experiments were organized using the intermediary service CloudResearch \citep{litman_turkprimecom_2017}, which allowed us to restrict participation to individuals in the United States. A small batch of participants (15) was recruited and played the game on December 12, 2022. The remainder of our participants were recruited and played the game on December 14, 2022. During play, human players do not have access to a log of past moves, but when the player makes a correct move a check mark is overlaid on the cleared piece (and the player can no longer click on the cleared piece). From this, a player can see the attributes of game pieces they have successfully cleared. Note that during processing of players' results, we discard instances where the player selects a piece but does not move it to a bucket; we call this situation a `finger-slip' and attribute it to a physical error made as the player tries to click and drag an object to a bucket. These finger-slips are thus not considered mistakes as we calculate $m^*$. 

\subsubsection{Experimental flow} 
Our experiments tasked players with playing 3 separate rules. As mentioned in the body of the manuscript, however, we only include results for the episodes of the first rule the player encounters (data from the second and third rules are not included in this manuscript). We plan to use experiments with multiple rules in later studies exploring the effects of transfer learning. Participants experienced the following order of events:
\begin{enumerate}
    \item Player is recruited via the Amazon Mechanical Turk platform.
    \item Player navigates to the link provided as part of the experiment and signs our participation consent form (see Appendix~\ref{app:consent}).
    \item Player receives instructions regarding the GOHR and the structure of the experiment.
    \item Player plays 3-5 episodes of rule 1. Note that after 3 episodes, the player has a persistent option to proceed to the next rule. After 3 episodes they also receive a persistent option to play 2 additional episodes (called `bonus rounds') as a chance to earn more points and a larger reward. As a result, players may play 3-7 episodes of this rule total, depending on their selections.
    \item Player proceeds to rule 2 and similarly plays 3-7 episodes of that rule (data not included in this manuscript).
    \item Player proceeds to rule 3 and similarly plays 3-7 episodes of that rule (data not included in this manuscript).
\end{enumerate}

\subsubsection{Subject counts}
Subject counts for each rule can be found in Table~\ref{tab:human-participants}. Note that our online interface requests that players provide a guess for the rule at the end of each episode. We found these guesses to be unreliable as measures of understanding, but we did use these guesses in tandem with performance as a filter for determining if a player made an honest effort during their time with the GOHR. We collected results from 24-26 participants for each starting rule (these players completed all 3 rules they were given). From this group we excluded 5 participants from our analysis. A player was only excluded if, for all 3 rules they played, they showed no effort to learn the rule (i.e., their error rates were consistent with guessing at random for all episodes and all rules) \textit{and} their inputted guesses did not relate to the game (e.g., if they provided only a random number so the button to proceed to the next episode would appear). Of the 5 excluded players, 1 was assigned to each of the following initial rules: SM1F, QN, QN2F, CWAF, BLTR.
\begin{table}[h]
\centering
\caption{Final count of human players for each rule in our experiments.}
\label{tab:human-participants}
\begin{tabular}{cc} \toprule
Rule & Count\\ \midrule
SM & 26\\
SM1F & 24\\
SM2O & 24 \\
CM & 25 \\
CM1F & 25 \\
CM2O & 26 \\
QN & 25 \\
QN2F & 24 \\
BLTR & 24 \\
BT & 26 \\
CW & 24 \\
CWAF & 24 \\
CW2F & 24 \\
\bottomrule
\end{tabular}
\end{table}

\subsubsection{Payment}\label{app:compensation}
Players were paid \$2.50-\$3.50, based on their performance, for approximately 20 minutes of time spent playing the GOHR. This amounts to an hourly wage of \$7.50-\$10.50. 

\subsubsection{Board generation parameters}
In order to reduce the likelihood that players receive drastically different boards, we enforced that each board must have at least one piece of each shape and color. Boards for stationary rules were generated with 9 pieces. Boards for non-stationary rules were generated with 8 pieces so that sequences would carry over smoothly from board to board (e.g., the Clockwise cycle would not be interrupted by episode transitions). We plan future investigations into boards with larger numbers of pieces as part of dedicated curricula studies. 

\subsubsection{IRB}\label{app:irb}
Our experiments with human players were reviewed and approved under the University of Wisconsin - Madison Minimal Risk Research IRB 2020-0781.

\subsubsection{Participation consent}\label{app:consent}
Prior to receiving instructions for the GOHR and beginning the experiment, players consented to participate in the study by clicking a checkbox at the conclusion of the following text:

``The task you are about to do is sponsored by University of Wisconsin-Madison. It is part of a protocol titled “Human and Machine Learning: The Search for Anomalies”. The purpose of this work is to compare reasoning biases in human and machine learners by testing what reasoning problems are relatively easier or more difficult for people, and for machines. More detailed instructions for this specific task will be provided on the next screen.

This task has no direct benefits. We do not anticipate any psychosocial risks. There is a risk of a confidentiality breach. Participants may become fatigued or frustrated due to the length of the study. The responses you submit as part of this task will be stored on a secure server and accessible only to researchers who have been approved by UW-Madison. Processed data with all identifiers removed could be used for future research studies or distributed to another investigator for future research studies without additional informed consent from the subject or the legally authorized representative. You are free to decline to participate, to end participation at any time for any reason, or to refuse to answer any individual question without penalty or loss of earned compensation. We will not retain data from partial responses. If you would like to withdraw your data after participating, you may send an email lupyan@wisc.edu or complete this form which will allow you to make a request anonymously. If you have any questions or concerns about this task please contact the principal investigator: Prof. Vicki Bier at vicki.bier@wisc.edu. If you are not satisfied with response of the research team, have more questions, or want to talk with someone about your rights as a research participant, you should contact University of Wisconsin’s Education Research and Social \& Behavioral Science IRB Office at 608-263-2320.

By clicking this box, I consent to participate in this task.''

\subsection{RL algorithms}\label{app:algos}
In this section we describe the details of our two sample RL players: DQN  and REINFORCE. These models come largely from canonical examples \citep{mnih_human-level_2015,sutton_simple_1992}, but are modified to include action masks tailored to fit the GOHR's mechanics.

\subsubsection{Experimental flow}
For all rules in our experiments, RL players received boards with 9 randomly generated game pieces. Given that RL players played many more episodes than humans, we did not similarly enforce that boards must have one game piece of each shape and color. Separate random seeds were used for each learning run (for both the game engine and the learning algorithm). As discussed in the body of the manuscript, each learning run consisted of random initialization of the model followed by serial play of a fixed number of episodes of the same rule.

\subsubsection{Feature maps}
As noted previously, to make the problem tractable we used feature mappings $\phi_i$ to represent the observed state $S\in\mathcal{S}$, i.e., $\phi_i(S)$ and maintain a static-sized input layer for the neural networks used in each algorithm. Given the plausible impact of different feature representations on task-based performance, we tested feature mappings that varied in the amount of memory included (i.e., the number of past boards and associated actions) as well as different representations of the past boards and actions themselves. The observed state may contain repeated board representations (i.e., $B_{t+1}=B_t$ if move $A_t$ is unsuccessful). Since the GOHR mechanics only update the rule state on successful moves, we only include distinct past board states and the associated successful action made from each board state in our feature maps. Repeated board states and the associated unsuccessful actions are not included.

Each featurization is the concatenation of a representation of the current board and a representation of some number of distinct past board states and associated actions. All featurizations used the same representation of the current board, defined as the concatenation of the following (assuming the use of the default set of four shapes and colors):
\begin{itemize}
    \item 4 36-long vectors, one corresponding to each shape. Each entry in the vector corresponds to one of the 36 cells on the board. An entry in the vector is 1 if a game piece of the corresponding shape is present in that cell and 0 otherwise.
    \item 4 36-long vectors, one corresponding to each color. Each entry in the vector corresponds to one of the 36 cells on the board. An entry in the vector is 1 if a game piece of the corresponding color is present in that cell and 0 otherwise. 
\end{itemize}
Totalling a 288-long boolean vector. The remainder of the featurization represents a chosen number of past board states and actions (2, 4, 6, or 8), each encoded `sparsely' or `densely' as follows.

A \textbf{sparse action} encoding is given by:
\begin{itemize}
    \item 1 144-long one-hot vector. Each entry corresponds to one of the 144 actions in the GOHR (i.e., placing a piece from one of the 36 cells into one of the 4 buckets). An entry in this vector is 1 for the index of the corresponding successful past action taken in that time step and 0 otherwise (i.e., only one entry will be 1).
\end{itemize}
A \textbf{dense action} encoding is given by the concatenation of the following:
\begin{itemize}
    \item 1 6-long one-hot vector. Each entry of the vector corresponds to one of the game board's 6 rows. An entry in this vector is 1 for the index of the row associated with the past successful action taken from this time step and 0 otherwise.
    \item 1 6-long one-hot vector. Each entry of the vector corresponds to one of the game board's 6 columns. An entry in this vector is 1 for the index of the column associated with the past successful action taken from this time step and 0 otherwise.
    \item 1 4-long one-hot vector. Each entry of the vector corresponds to one of the game board's 4 buckets. An entry in this vector is 1 for the index of the bucket associated with the past successful action taken from this time step and 0 otherwise.
\end{itemize}
Totalling a 16-long boolean vector (with exactly 3 1's and the remainder 0).

A \textbf{sparse board} encoding is the 288-long representation of the entire past board, following the same encoding procedure as the current board. Given that board states necessarily change by only one game piece at a time, we define a \textbf{dense board} encoding to be the concatenation of:
\begin{itemize}
    \item 1 4-long vector, corresponding to the length of the list of shapes. The vector is one-hot encoded corresponding to the index of the shape removed from the board in that time step.
    \item 1 4-long vector, corresponding to the length of the list of colors. The vector is one-hot encoded corresponding to the index of the color removed from the board in that time step.
\end{itemize}
Totalling an 8-long boolean vector (with exactly 2 1's and the remainder 0). Given the board information in the present time step and the action information, a model with a dense board encoding still retains the information needed to reconstruct complete board states.

Thus, we can summarize the five types of feature maps used in our testing, where $n$ is the number of steps of memory:
\begin{enumerate}
    \item Board dense, action dense (\textbf{BD-AD}): The standard representation of the current board and dense representations of the past $n$ board states and successful actions. Total of $288+(8+16)n$ boolean inputs.
    \item Board dense, action sparse (\textbf{BD-AS}): The standard representation of the current board, dense representation of the past $n$ board states, and sparse representations of the past $n$ successful actions. Total of $288+(8+144)n$ boolean inputs.
    \item Board sparse, action dense (\textbf{BS-AD}): The standard representation of the current board, sparse representations of the past $n$ board states, and dense representations of the past $n$ successful actions. Total of $288+(288+16)n$ boolean inputs.
    \item Board sparse, action sparse (\textbf{BS-AS}): The standard representation of the current board and sparse representations of the past $n$ board states and successful actions. Total of $288+(288+144)n$ boolean inputs.
    \item Board sparse-dense, action sparse-dense (\textbf{BSD-ASD}): The standard representation of the current board and concatenations of both sparse and dense representations of the past $n$ board states and actions. Total of $288+(288+8+144+16)n$ boolean inputs.
\end{enumerate}
We tested each of these feature maps with $n=2,4,6,8$ steps of memory, corresponding to 20 distinct featurizations. Featurizations range from 336 boolean inputs (BD-AD, $n=2$) to 3936 boolean inputs (BSD-ASD, $n=8$).

\subsubsection{DQN}
Our DQN player, with algorithm pseudocode shown in Algorithm~\ref{alg:dqn}, follows the canonical DQN structure as presented in \citet{mnih_human-level_2015}, modified to include an action mask. First, as described in \cite{van_hasselt_deep_2016}, the algorithm reduces value-to-go overestimation by maintaining a policy network (parameterized by $\theta^p$) and a target network (parameterized by $\theta^t$) which are periodically synchronized (every $T$ iterations). The next action $a'$ used for calculation of the TD-0 target, $y$, is chosen as the argmax over the policy network, but the $Q$ value used to calculate the TD-0 estimate uses the target network. Second, we employ an action mask that prevents the algorithm from selecting actions that do not correspond to items currently on the game board; that is, the valid action set $\hat A_t(\phi(S_t)) \subset \mathcal{A}$ is the subset of action indices associated with objects currently present on the board per $\phi(S_t)$. Given that the rule state does not change on incorrect moves, we also maintain a list of incorrect moves made since the last successful move and these actions are also included in the action mask (in order to prevent the algorithm from repeatedly making the same mistake until the next successful move). In practice, we achieve this action mask by modifying the Q-value of invalid actions in the output layer to be less than that of any valid action. When moves are selected randomly, we choose uniformly from the set of valid actions. The algorithm makes random actions with probability $\epsilon$, which decays exponentially (updated each move) from $\epsilon_{start}$ to $\epsilon_{end}$ per a move-based decay rate $\epsilon_{decay}$, i.e., $\epsilon = \epsilon_{end} + (\epsilon_{start}-\epsilon_{end})\exp(\nicefrac{m}{\epsilon_{decay}})$, where $m$ is the move index of the entire learning run. Greedy actions are taken with probability $1-\epsilon$. Note that our algorithm includes an intra-episode move limit ($L$, if this is reached, the episode concludes), but given the action mask and that boards only contained 9 pieces, this limit was not reached during our experiments. In practice, the algorithm is implemented using the PyTorch library \citep{paszke_pytorch_2019}.

\floatname{algorithm}{Algorithm}
\begin{algorithm}
\textbf{Input} : Input featurization method $\phi$, sync count $T$, episode count $M$, buffer size $N$, batch size $B$, episode move limit $L$.
\begin{algorithmic}
    \State Initialize experience replay buffer $\mathcal{R}$ with capacity $N$.
    \State Initialize the network parameters  $\theta^{p}$ with random weights, set $\theta^{t}\leftarrow \theta^{p}$
    \For{episode $e = 0,\dots,M$} 
    \State Draw initial state $s_0$ (i.e., board state $b_0$) from the game engine, map to $\phi(s_0)$
    \For {time step $t=0,\dots,L$ (or until board cleared)} 
    \State Select valid action $a_t$, randomly with probability $\epsilon$, otherwise $a_t=$argmax$_aQ(\phi(s_t),a;\theta^p)$
    \State Take action $a_t$ and observe reward $r_t$ and board state $b_{t+1}$.
    \State Set $s_{t+1} \leftarrow (s_t,a_t, b_{t+1})$ and add $\left(\phi(s_t),a_t,r_t,\phi(s_{t+1})\right)$ to $\mathcal{R}$.
    \State If $t>B$, randomly sample a batch $\{\left(\phi(s_j),a_j,r_j,\phi(s'_{j})\right)\}|_{j=1}^B$ from $\mathcal{R}$, 
    \State \quad and set $y_j = r_j + (1-\mathbf{1}_{\{cleared\}})\gamma \max_{a'} Q(\phi(s'_{j}), a'; \theta^{t})$
    \State Perform optimization step on Huber loss $\mathcal{L}(Q(\phi(s_j),a_j;\theta^p)-y_j)$
    \State Every $T$ steps set $\theta^t\leftarrow \theta^p$
    \EndFor
    \EndFor
\end{algorithmic}
\caption{DQN with Experience Replay}
\label{alg:dqn}
\end{algorithm}

\subsubsection{REINFORCE}
Our REINFORCE player, with algorithm pseudocode shown in Algorithm~\ref{alg:rn}, resembles the canonical REINFORCE algorithm with baseline given by \citet{sutton_simple_1992}, modified to include an action mask and using the average trajectory return as the baseline. The learner's policy is approximated with a neural network, parameterized by $\theta$. Similar to DQN, the action mask for REINFORCE prevents the algorithm from selecting actions that do not correspond to items currently on the game board; that is, the valid action set $\hat A_t(\phi(S_t)) \subset \mathcal{A}$ is the subset of action indices associated with objects currently present on the board per $\phi(S_t)$. Actions are sampled from this subset using a softmax operation on the output layer of the neural network. In practice, this mask is achieved by writing the network outputs associated with invalid actions to $-\infty$ prior to application of the softmax operation over all 144 actions. Note that, unlike DQN, incorrect moves made since the last correct move were not included in the REINFORCE action mask; REINFORCE samples actions from the masked action distribution, making it less likely to get stuck on an incorrect move than DQN (especially once the exploration parameter has decayed significantly). As with DQN, we implement our REINFORCE player using the PyTorch library \citep{paszke_pytorch_2019}.

\floatname{algorithm}{Algorithm}
\begin{algorithm}
\textbf{Input} : Input featurization method $\phi$, episode count $M$, episode move limit $L$, learning rate $\alpha$.
\begin{algorithmic}
    \State Initialize the network parameters  $\theta$ with random weights
    \For{episode $e = 0,\dots,M$} 
    \State Draw initial state $s_0$ (i.e., board state $b_0$) from the game engine, map to $\phi(s_0)$
    \State Initialize record of episode trajectory $\mathcal{T}$, reset $L$ to inputted value
    \For {time step $t=0,\dots,L$} 
    \State Sample valid action $a_t$ per policy $\pi(\phi(s_t))$
    \State Take action $a_t$ and observe reward $r_t$ and board state $b_{t+1}$
    \State Append $(\phi(s_t),a_t,r_t)$ to $\mathcal{T}$
    \If {board cleared}
    \State $L=t$
    \EndIf
    \EndFor
    \State Calculate trajectory returns $G_t\leftarrow\sum_{k=t}^{T}\gamma^{k-t}r_k$, $t=0,\dots,L$
    \State Calculate baseline $\hat{v} \leftarrow \frac{1}{L+1}\sum_{k=0}^{L} G_k$
    \For {time step $t=0,\dots,L$}
    \State $\delta \leftarrow G_t-\hat{v}$
    \State Perform gradient step $\theta \leftarrow \theta + \alpha \delta \nabla \ln \pi(a_t|\phi(s_t);\theta)$
    \EndFor
    \EndFor
\end{algorithmic}
\caption{REINFORCE}
\label{alg:rn}
\end{algorithm}

\subsubsection{Hyperparameter selections}\label{app:alg_hyp}
We note that under our reward scheme (immediate feedback of $0$ or $-1$), the mechanics of the GOHR lend themselves to a model with strong discounting (i.e., discount factor $\gamma$ near 0). Both DQN and REINFORCE used $\gamma=0.001$. We used Optuna \citep{akiba_optuna_2019} to perform a coarse hyperparameter tuning for both DQN and REINFORCE, after which we fine tuned parameters manually. Chosen parameters are not necessarily optimal for any given rule, but represent a tuning which performs well across all tested rules. Our focus for our sample experiments was to explore task structure and input featurization methods; the GOHR enables a similarly granular study of performance with respect to hyperparameter selections, but this was not our focus in this paper. Hyperparameters for our DQN and REINFORCE models can be found in Table~\ref{tab:hyper}.

\begin{table}[h!]
\centering
\caption{DQN and REINFORCE hyperparameters. Hyperparameters that are irrelevant for a given model are marked `-'.}
\label{tab:hyper}
\begin{tabular}{ccc} \toprule
Hyperparameter & DQN value & REINFORCE value  \\ \midrule
Discount factor $\gamma$ & 0.001 & 0.001 \\
Activation function & ReLU  & LeakyReLU\\
Hidden layer count & 2 & 1\\
Hidden layer size & 700 & 1000\\
Replay memory size $N$ & 30000 & - \\
Learning batch size $B$ & 256 & -\\
Learning rate $\alpha$ & 0.0001 & 0.0007\\
$\epsilon$ start & 0.99 & -\\
$\epsilon$ end & 0.0001 & -\\
$\epsilon$ decay rate & -200 & -\\
Sync count $T$ & 100 & -\\
Episode move limit $L$ & 100 & 100 \\
Optimizer & RMSprop & RMSprop\\
\bottomrule
\end{tabular}
\end{table}

\subsubsection{Convergence criteria}\label{app:algos_metrics}
To account for algorithm stochasticity and give a fair comparison to humans (for whom a background rate of errors was expected and observed), we allow a deviation from error-free play for our RL players. We define a convergent learning run to be a move-based error rate of less than $0.0025$ over the final 150 episodes of play. Given that boards have 9 pieces each, this amounts to the algorithm making 3 or fewer errors in the final 150 episodes of play. We note that we saw much faster convergence behavior for DQN over REINFORCE. This was likely due to a number of factors, such as the stochasticity in REINFORCE's action selection and the fact that DQN performs training updates each move rather than following each episode (and benefits from past experiences through its experience replay buffer). While shape- and color-based rules showed the slowest convergence (with some runs failing to meet our criteria), play on most rules converged well before our chosen episode limits. Ultimately, such a threshold will be inherently arbitrary; if the algorithms were permitted additional episodes of play, we could have enforced stricter convergence criteria. However, for the purposes of our sample experiments, we believe these conditions well-characterize the learning performance of our algorithms relative to humans and little insight would be gained from additional playtime (and associated stricter convergence conditions).

\subsubsection{Compute resources}\label{app:alg_compute}
All experiments were performed on the University of Wisconsin-Madison's Center for High Throughput Computing cluster. Jobs were batched such that each learning run received 1 compute core (non-standardized hardware), 0.25-1.25GB of RAM, and 100MB of disk space. Learning run durations differed based on algorithm and featurization choice (along with variations in cluster resource availability), but each run generally completed in 0.5-3 hours. No GPU resources were used in our experiments.

\subsection{Supplementary results}\label{app:addl_plots}
\subsubsection{Color-based rule results}\label{app:color_rules}
\begin{figure}[h]
    \centering
    \includegraphics[width=.6\textwidth]{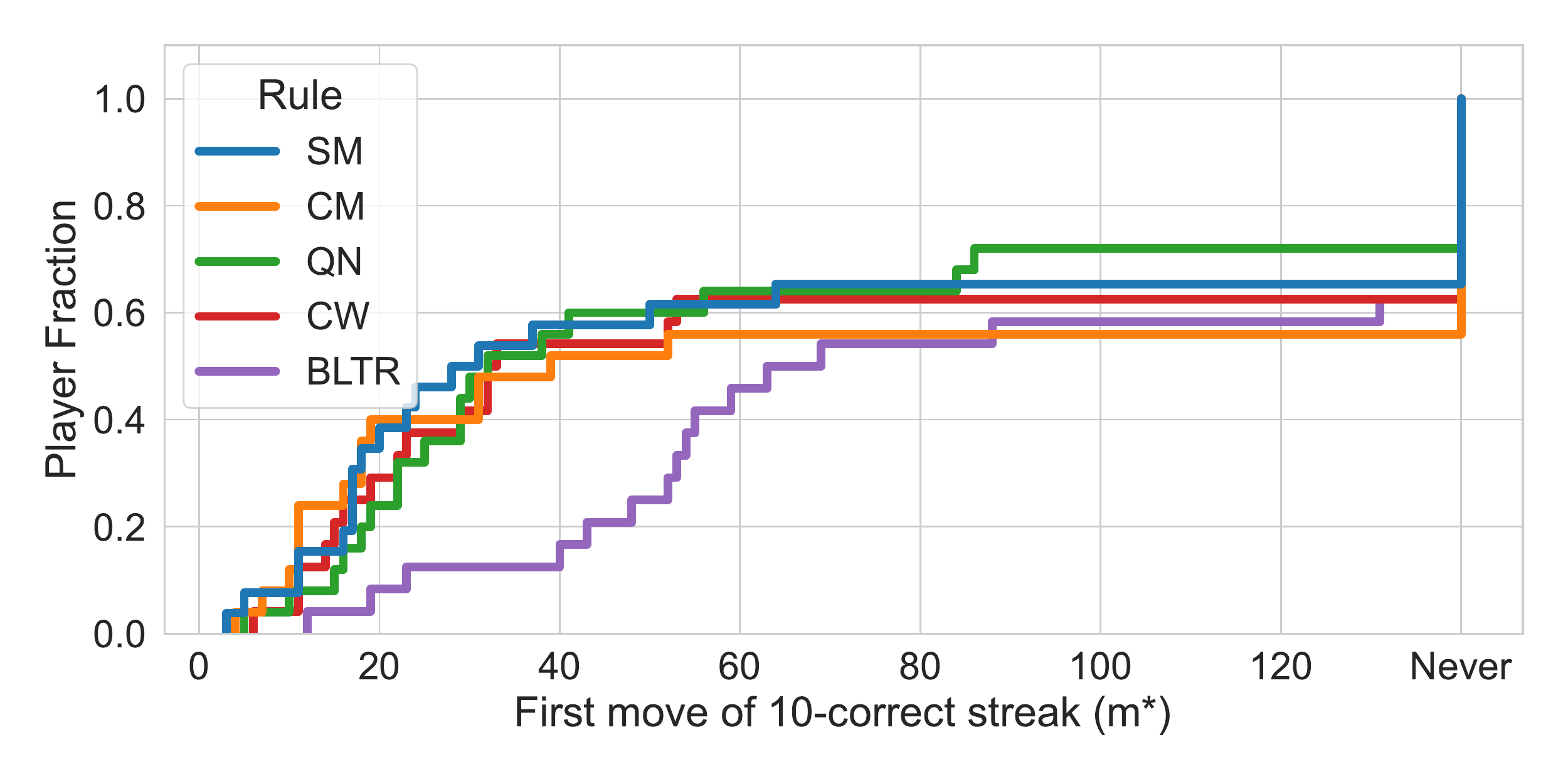}
    \caption{Empirical cumulative distribution plots denoting the fraction of players who achieved an $m^*$ streak by a given move index for all base rules (including CM).}
    \label{fig:human_ecdf_basewithcolor}
\end{figure}

\begin{figure}[h]
    \centering
    \includegraphics[width=.6\textwidth]{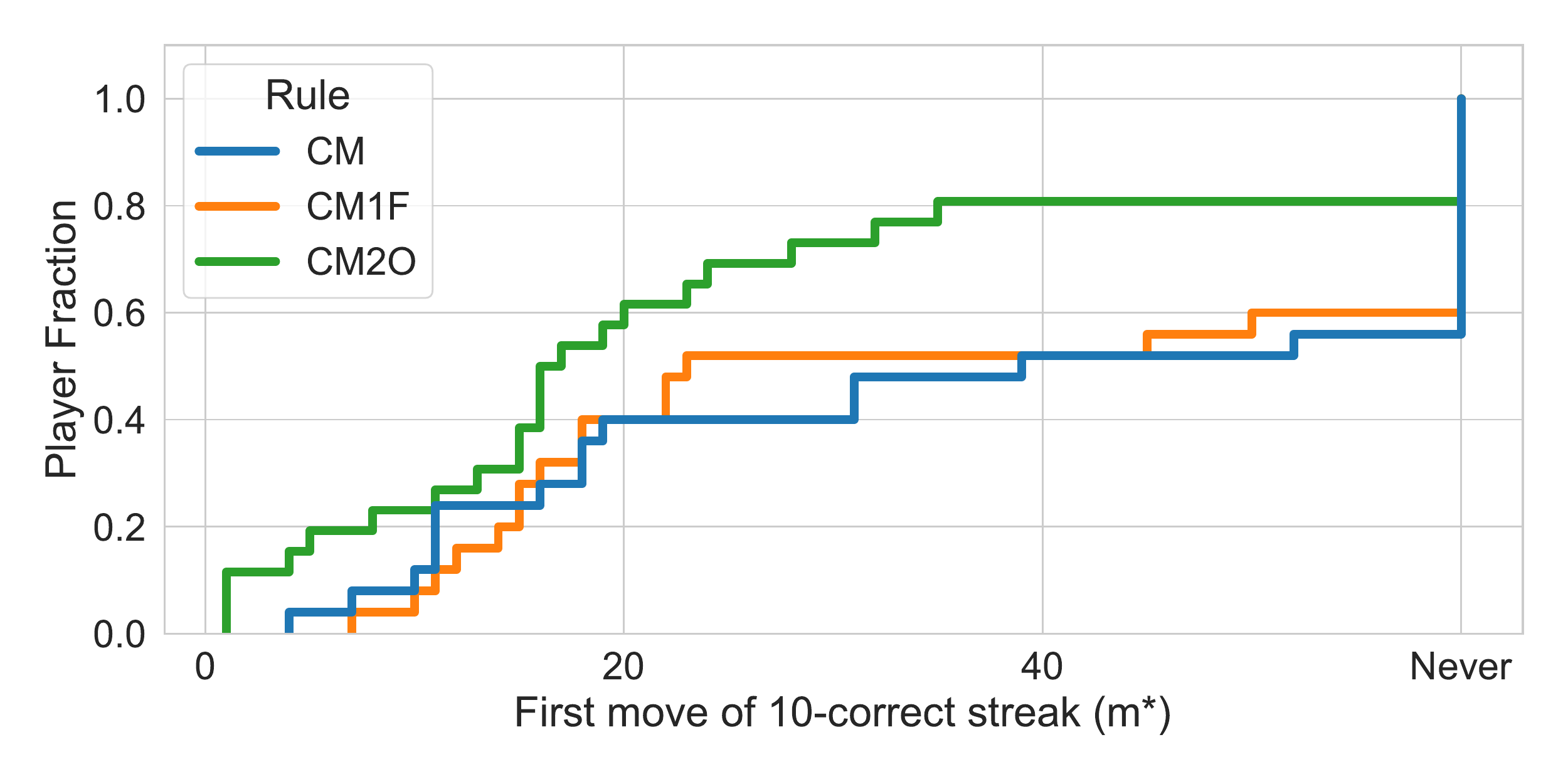}
    \caption{Empirical cumulative distribution plots denoting the fraction of players who achieved an $m^*$ streak by a given move index for rules in the color family.}
    \label{fig:human_ecdf_color}
\end{figure}

Human results for the Color Match (CM) rule, in context with the other base rules from our first experiment, can be found in Figure~\ref{fig:human_ecdf_basewithcolor}. Human results for the complete color-based generality family (CM, CM1F, CM2O) can be found in Figure~\ref{fig:human_ecdf_color}. As noted in the body of the manuscript, these resemble the results from the shape family.

\subsubsection{Rule generality RL results}\label{app:rl_rule_gen}
\begin{figure}[h]
    \centering
    \includegraphics[width=.8\textwidth]{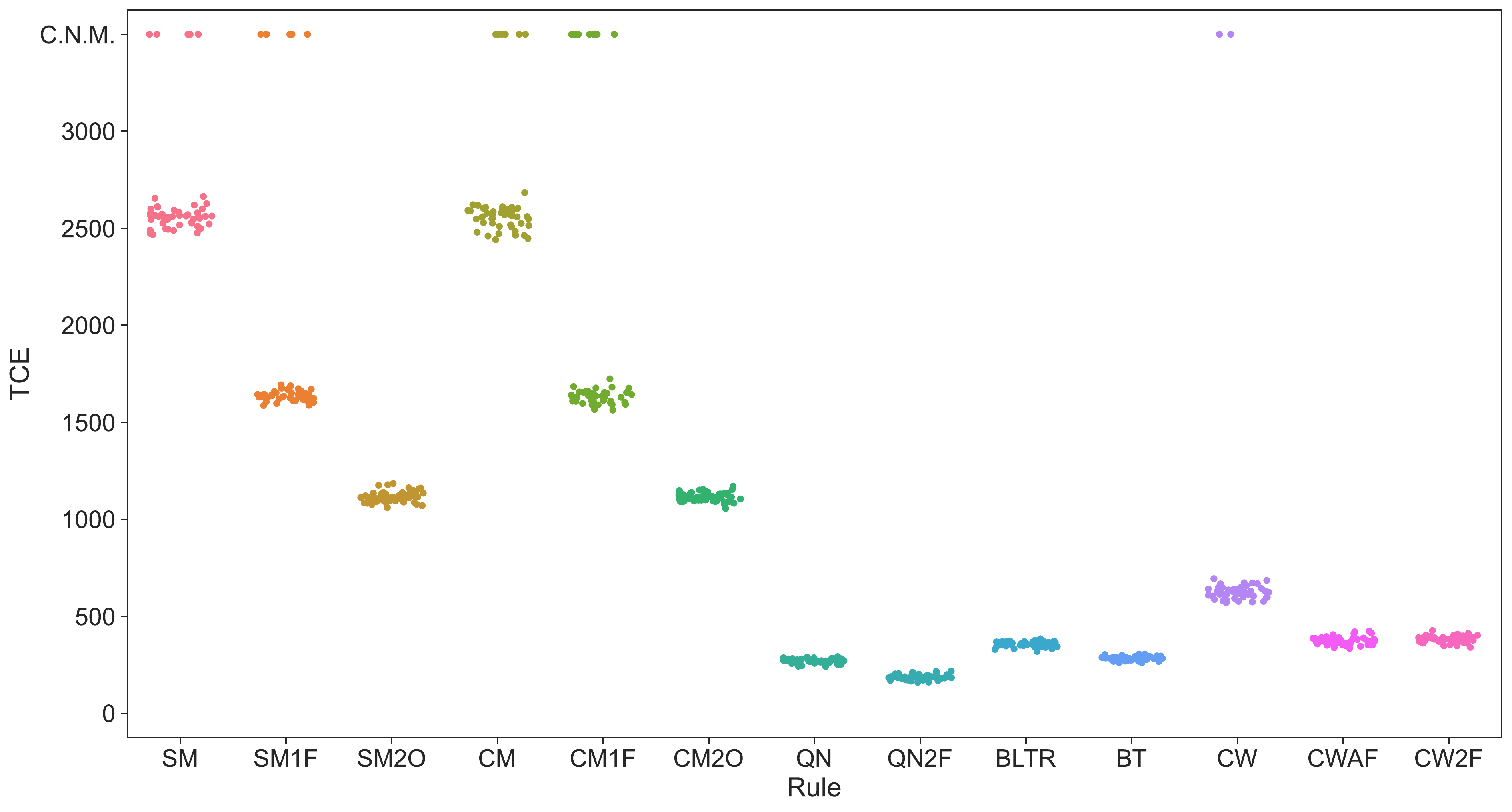}
    \caption{Strip plots for DQN TCE distributions across all rules. Each dot represents a learning run for the corresponding rule. `C.N.M.' denotes that convergence criteria were not met for that learning run.}
    \label{fig:rl_gen_dqn}
\end{figure}

\begin{figure}[h]
    \centering
    \includegraphics[width=.8\textwidth]{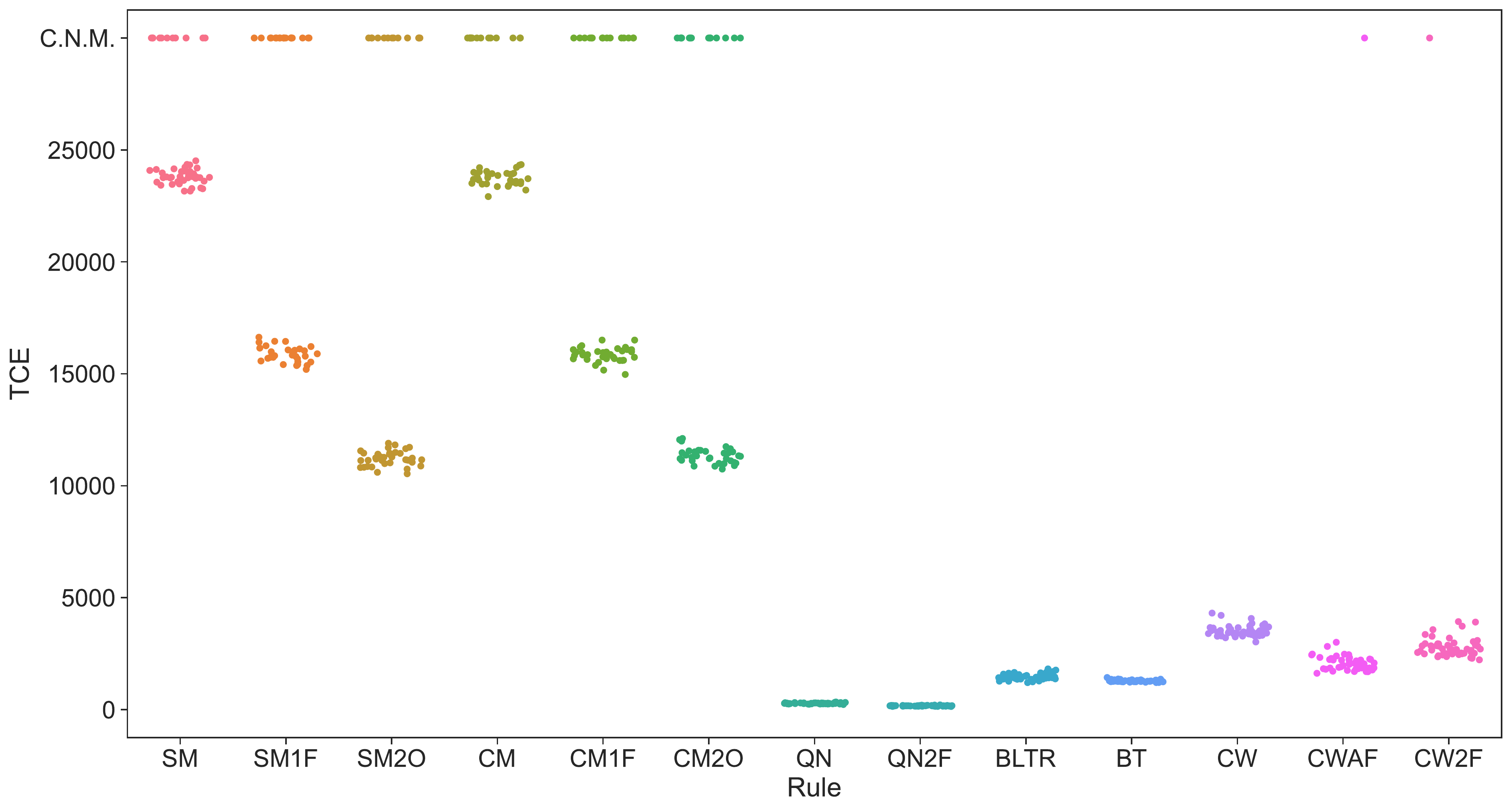}
    \caption{Strip plots for REINFORCE TCE distributions across all rules. Each dot represents a learning run for the corresponding rule. `C.N.M.' denotes that convergence criteria were not met for that learning run.}
    \label{fig:rl_gen_rn}
\end{figure}

Strip plots of TCE distributions for all rules involved in the generality comparisons included in our second experiment can be found in Figure~\ref{fig:rl_gen_dqn} (DQN) and Figure~\ref{fig:rl_gen_rn} (REINFORCE). As noted in the body of the manuscript, we see uniformly better performance for more general rule variants (though we note that even longer runs would be needed for all shape/color rule learning runs to meet our convergence criteria, particularly for REINFORCE).
\subsubsection{Base rule comparisons (all players)}\label{app:base_rule_compare}
As noted in the body of the manuscript, for our first experiment we gathered the $m^*$ and TCE distributions for all base rules and performed all possible pairwise rule comparisons (separately for each learner). Table~\ref{tab:base-stat-tests} shows the results of two-sided Mann-Whitney U-Tests for all rule pairs and learners (human, DQN, and REINFORCE), including Color Match (CM). Note that the DQN and REINFORCE results are grouped together for visual compactness (no comparisons were done between the TCE distributions for DQN and REINFORCE). We see one rule pair with significantly different performance for humans (QN-BLTR). All rule pairings, except SM-CM, show significantly different performance for both RL algorithms. Note that given the equivalent logical structure of SM and CM and their equivalent treatment in our featurizations, we would expect p-values for this comparison of approximately 0.5. We see values of 0.624 and 0.908 for the DQN and REINFORCE U-tests, respectively. This deviation is due to chance differences in the number of learning runs which meet our convergence criteria. Since non-convergent runs are assigned placeholder values of TCE higher than all convergent runs, comparisons of these distributions are affected by the number of convergent runs for each. REINFORCE, in particular, would have needed an impractical number of additional episodes of play to meet convergence criteria for all runs. As a result, the strip plots in Figure~\ref{fig:rl_gen_dqn} and Figure~\ref{fig:rl_gen_rn} provide a better view of the effectively identical performance of the RL algorithms on SM and CM, as we would expect.
\begin{table}[h]
\centering
\caption{P-value results of two-sided U-Tests comparing performance on the base rules presented in our first experiment. Each entry of the table compares the rule on the left to the rule given above (i.e., top-left entry compares SM to CM). Results are separated for humans (left) and RL algorithms (right). Tests use $m^*$ and TCE as point metrics for HL and RL, respectively. Values <0.001 are denoted by * and significant results at the $\alpha=0.05$ level are highlighted with boldface. For RL players, significance results are given in the form (p-value for DQN / p-value for REINFORCE).}
\label{tab:base-stat-tests}
\begin{tabular}{ccccccccc} \toprule
& \multicolumn{4}{c}{Human} & \multicolumn{4}{c}{RL (DQN/REINFORCE)} \\ \cmidrule(r){2-5} \cmidrule(r){6-9}
Rule  & CM & QN & BLTR & CW & CM & QN & BLTR & CW \\ \midrule
SM & 0.793 & 0.709 & 0.055 & 0.720 & 0.624/0.908  & \textbf{*}/\textbf{*} & \textbf{*}/\textbf{*} & \textbf{*}/\textbf{*} \\
CM  & - &  0.952 & 0.178 & 0.892 & -  & \textbf{*}/\textbf{*} & \textbf{*}/\textbf{*} & \textbf{*}/\textbf{*} \\
QN  & - &  - & \textbf{0.049} & 0.927 & -  & - & \textbf{*}/\textbf{*} & \textbf{*}/\textbf{*} \\
BLTR  & - & - & - & 0.102 & -  & - & - & \textbf{*}/\textbf{*}  \\\bottomrule
\end{tabular}
\end{table}

\subsection{Featurization discussion}\label{app:feat_discussion}
\begin{figure}[h]
    \centering
    \includegraphics[width=\textwidth]{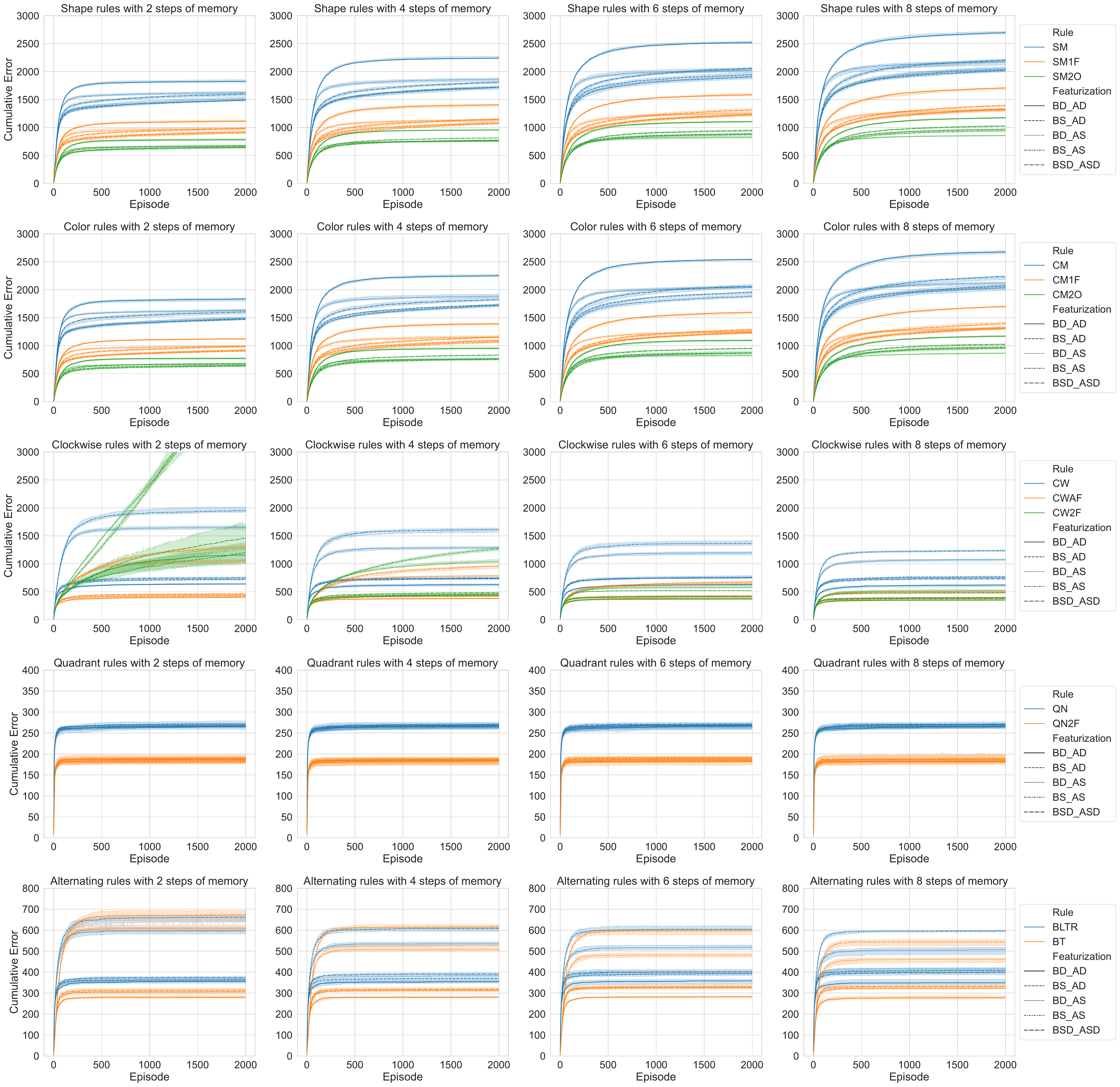}
    \caption{DQN median learning curves, plotting median cumulative error versus episode. Each line summarizes the median performance of 20 learning runs for a specific rule (noted by the line color) using a specific feature map (noted by the line style). Each row of plots has the same y-axis scaling and considers the same set of rules. Each column of plots considers featurizations with a different selection of memory. Shaded regions denote 95\% confidence intervals obtained from 500 bootstraps.}
    \label{fig:dqn_med_learn}
\end{figure}

\begin{figure}[h]
    \centering
    \includegraphics[width=\textwidth]{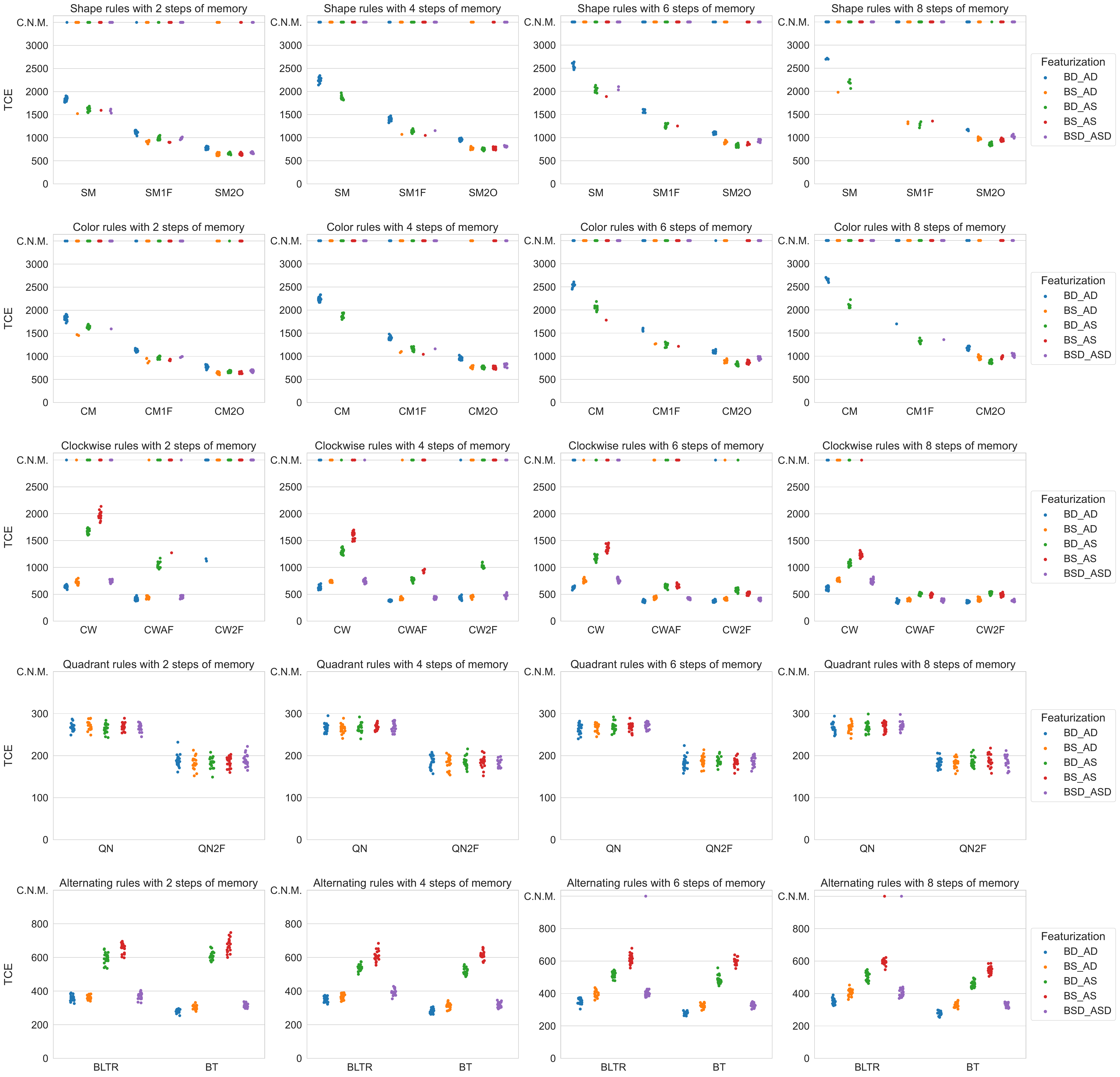}
    \caption{DQN TCE strip plots, where each dot summarizes the TCE of a particular learning run (TCE given on the y-axis and corresponding rule given on the x-axis). Each row of plots has the same y-axis scaling and considers the same set of rules. Each column of plots considers featurizations with a different selection of memory. With memory fixed within each plot, dot color denotes the feature map used in that learning run. Each plot contains 20 learning runs for each rule-feature map pair. `C.N.M.' denotes that the convergence criteria were not met for that learning run.}
    \label{fig:dqn_term_strip}
\end{figure}

\begin{figure}[h]
    \centering
    \includegraphics[width=\textwidth]{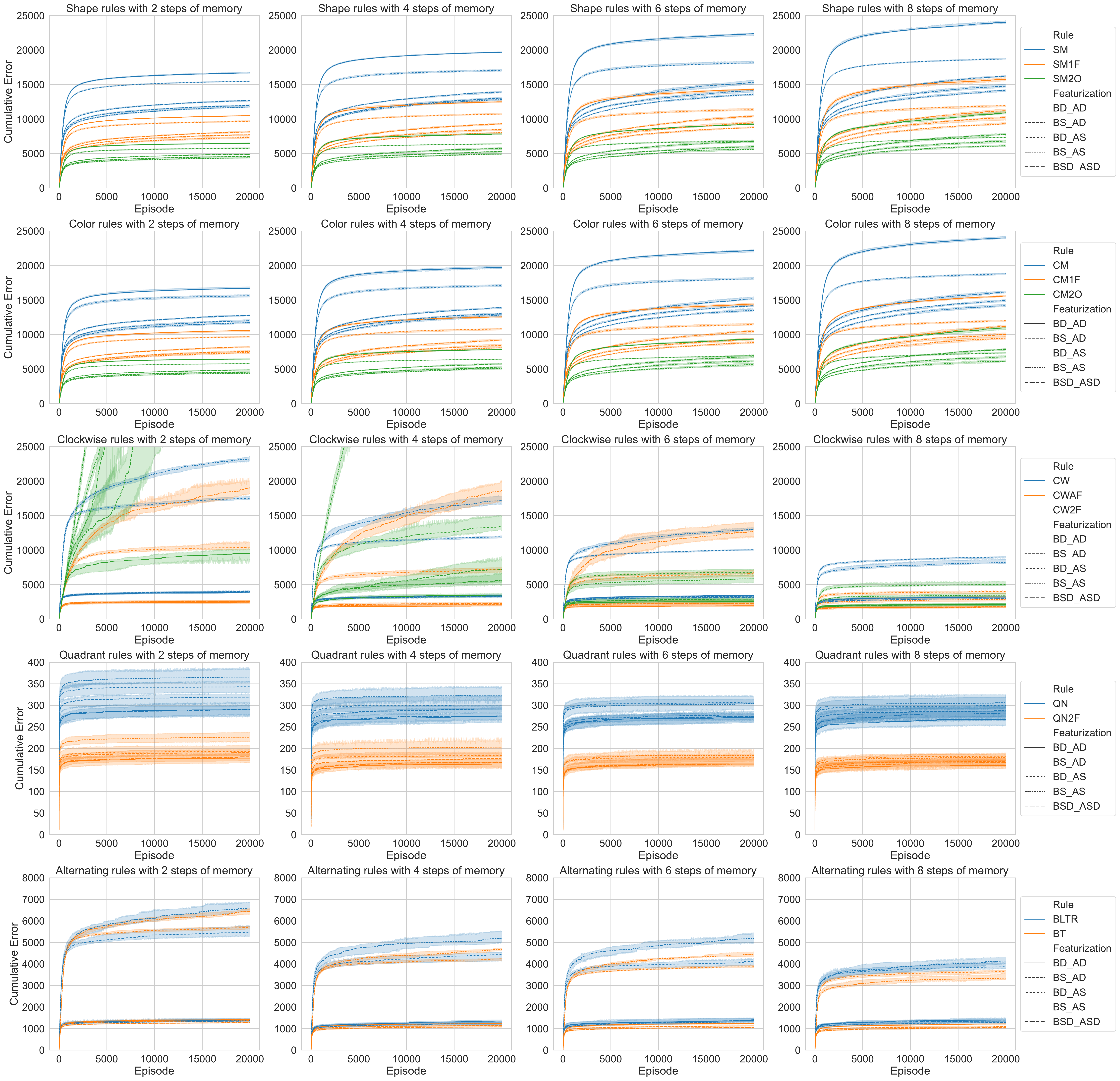}
    \caption{REINFORCE median learning curves, plotting median cumulative error versus episode. Each line summarizes the median performance of 20 learning runs for a specific rule (noted by the line color) using a specific feature map (noted by the line style). Each row of plots has the same y-axis scaling and considers the same set of rules. Each column of plots considers featurizations with a different selection of memory. Shaded regions denote 95\% confidence intervals obtained from 500 bootstraps.}
    \label{fig:rn_med_learn}
\end{figure}

\begin{figure}[h]
    \centering
    \includegraphics[width=\textwidth]{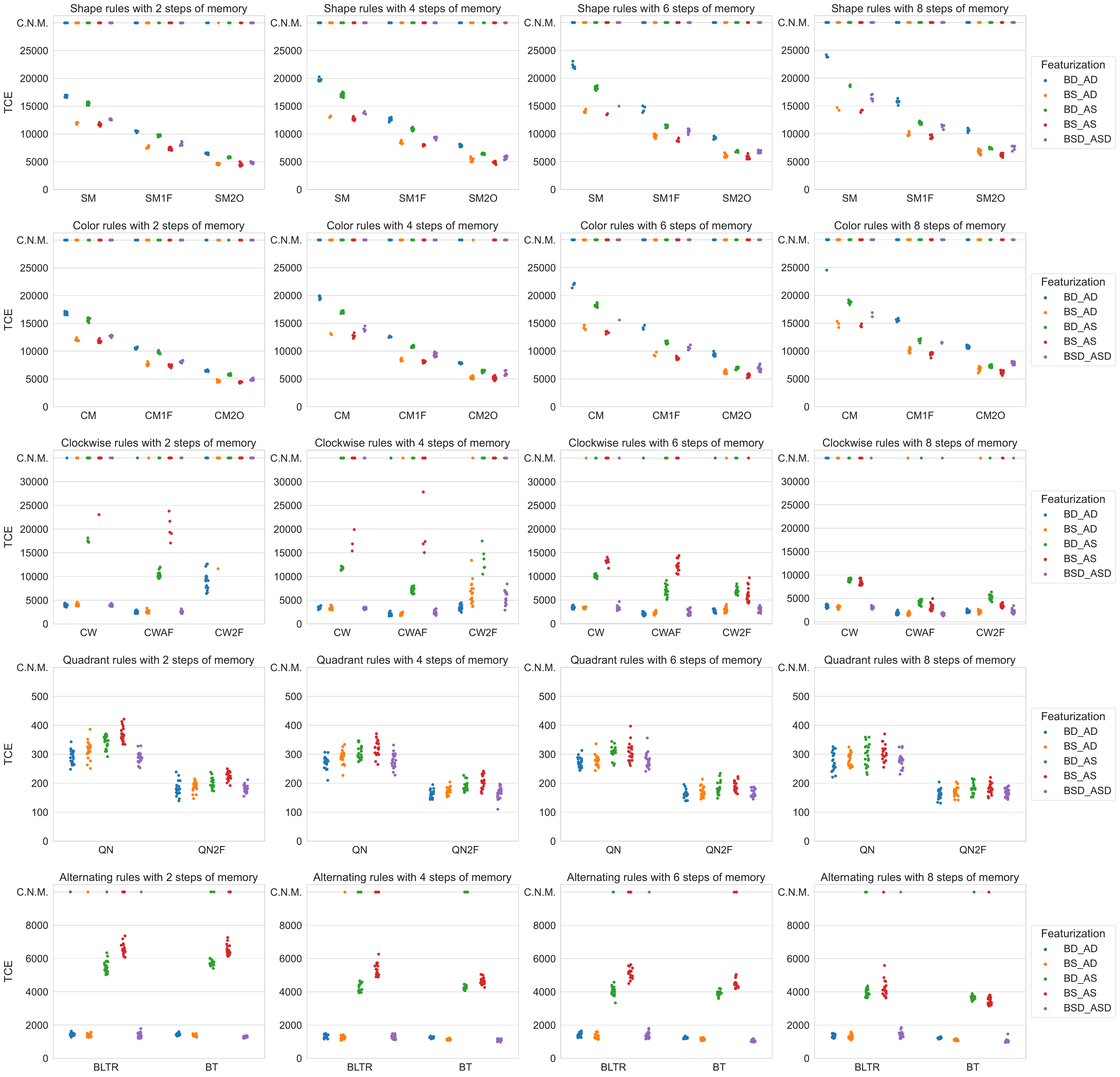}
    \caption{REINFORCE TCE strip plots, where each dot summarizes the TCE of a particular learning run (TCE given on the y-axis and corresponding rule given on the x-axis). Each row of plots has the same y-axis scaling and considers the same set of rules. Each column of plots considers featurizations with a different selection of memory. With memory fixed within each plot, dot color denotes the feature map used in that learning run. Each plot contains 20 learning runs for each rule-feature map pair. `C.N.M.' denotes that the convergence criteria were not met for that learning run.}
    \label{fig:rn_term_strip}
\end{figure}

As noted in Appendix~\ref{app:algos}, our experiments included testing with 20 distinct input featurization methods, resulting from all combinations of our five feature maps (BD-AD, BS-AD, BD-AS, BS-AS, BSD-ASD) and four levels of memory (2 steps, 4 steps, 6 steps, 8 steps). We found that no featurization outperformed all others across all tested rules; different choices of feature map and memory led to tradeoffs in performance. We found two types of plots helpful in interpreting the effects of different featurizations: line plots of episodic median performance and strip plots of TCE. Plots showing episodic median performance gather the median cumulative error value at each episode across all learning runs and are useful for comparing convergence differences between rule-featurization pairs. Strip plots of TCE capture end-of-run behavior more succinctly, but lose resolution in describing convergence behavior (in particular among runs which do not meet our convergence criteria). To observe performance trends, we create both types of plots using a grid-style layout. Each row of plots corresponds to a family of rules (e.g., shape rules SM, SM1F, and SM2O) and each column corresponds to a number of steps of memory (2, 4, 6, or 8). Each plot within the grid shows performance (either line plots of median cumulative error or TCE strip plots) for all rules in that row's rule family under all feature maps using that column's amount of memory. Figure~\ref{fig:dqn_med_learn} shows DQN episodic median performance and Figure~\ref{fig:dqn_term_strip} shows strip plots of DQN TCE. Figures~\ref{fig:rn_med_learn} and \ref{fig:rn_term_strip} similarly show REINFORCE episodic mean performance and strip plots of TCE, respectively. These plots gather data from 20 learning runs per featurization-rule pair (for each algorithm). Note that we use abbreviated episode horizons (2000 for DQN and 20000 for REINFORCE) to reduce the computational burden of these experiments; the abbreviated horizons are sufficient to show relevant performance trends. We note that although we see large differences between DQN and REINFORCE in performance (measured by TCE), both algorithms show similar performance trends with respect to tested featurizations. As such, we discuss these broad trends jointly rather than for DQN and REINFORCE individually.

As noted in the body of the manuscript, increasing the amount of memory in a featurization generally improves performance on non-stationary rules. The rule CW2F highlights how an algorithm must have sufficient memory to distinguish between meaningfully different observations (i.e., the rule must be identifiable); in CW2F two moves of the 0-1-*-* sequence are free, so it is important that the model have more than 2 steps of memory in order to correctly learn from observed feedback. Gains of increasing memory appear to be diminishing, however, as CW2F illustrates. While we note a large improvement across feature maps from $n=2$ to $n=4$ steps of memory on CW2F, moving to $n=6$ and $n=8$ shows decreasing marginal performance gains. We see even earlier diminishing returns to increased memory, as expected, for rules which are identifiable with fewer steps of memory (such as other clockwise rules or the alternating rules). Unfortunately, the performance gains on non-stationary rules from additional memory generally correspond to worsened performance on stationary rules. In particular, the shape and color families show progressively worse performance as memory is added to the input featurization. Interestingly, performance on QN and QN2F remains essentially static with respect to choice of memory. This may suggest that, in contrast to the shape or color rules, the quadrant rules are so easily learned from the representation of the current board that the models are unaffected by the addition of largely useless historical board/action information. In general, if the structure of the learning tasks an algorithm might encounter is not known a priori, the algorithm designer may need to choose an amount of memory based on their tolerance for the potential performance tradeoffs on tasks of different structure.

We also note interesting behavior among the five choices of feature maps (BD-AD, BS-AD, BD-AS, BS-AS, BSD-ASD) across our tested rules. In general, sparse board representations perform more poorly than dense representations (i.e., BD-AD tends to outperform BS-AD and BD-AS tends to outperform BS-AS). This follows our intuition, as the large number of additional features included in sparse board representations are generally not useful for our tested rules; we might note different behavior for rules with a stronger dependence on complete past board arrangements. Action representations, however, appear to bring more explicit performance tradeoffs for our tested rules. Dense action representations make the specific row, column, and bucket selection of a past move more readily available than sparse action representations. As might be expected, this improves performance on non-stationary tasks that explicitly depend on past bucket information, such as the clockwise and alternating families. For the color- and shape-based rules, though, we see some benefits to using sparse action representations over dense representations; BD-AS, for example, appears to show the most consistent convergence behavior across all featurizations for the shape/color rules, even as memory is increased. Performance on the stationary quadrant-based rules appears invariant to the chosen feature map, as was also the case with respect to choice of memory. Again, we believe this is because the quadrant rules are so easily learned from the representation of the current board that performance is unaffected by changes to the representation of historical information. We expect future work investigating a broader set of task structures to elucidate the situational value of different feature maps. Importantly, we note that the combined feature map BSD-ASD, which contains both sparse and dense representations, does not always resemble the best performing feature map (among BD-AD, BS-AD, BD-AS, BS-AS). While BSD-ASD performs well (though not best) for clockwise, quadrant, and alternating rules, it shows poor performance on the shape and color rules, similar to other feature maps with sparse board representations. While these results do not suggest a simple answer regarding featurization methods, they do highlight the importance of considering the structure of tasks that an algorithm might encounter when designing an algorithm's input featurization.

\end{document}